\PassOptionsToPackage{unicode}{hyperref}
\PassOptionsToPackage{hyphens}{url}
\PassOptionsToPackage{dvipsnames,svgnames,x11names}{xcolor}
\documentclass[
  12pt]{article}

\usepackage{amsmath,amssymb}
\usepackage{iftex}
\ifPDFTeX
  \usepackage[T1]{fontenc}
  \usepackage[utf8]{inputenc}
  \usepackage{textcomp}
\else 
  \usepackage{unicode-math}
  \defaultfontfeatures{Scale=MatchLowercase}
  \defaultfontfeatures[\rmfamily]{Ligatures=TeX,Scale=1}
\fi
\usepackage{lmodern}
\ifPDFTeX\else  

\fi
 
\IfFileExists{upquote.sty}{\usepackage{upquote}}{}
\IfFileExists{microtype.sty}{
  \usepackage[]{microtype}
  \UseMicrotypeSet[protrusion]{basicmath} 
}{}
\makeatletter
\@ifundefined{KOMAClassName}{
  \IfFileExists{parskip.sty}{%
    \usepackage{parskip}
  }{
    \setlength{\parindent}{0pt}
    \setlength{\parskip}{6pt plus 2pt minus 1pt}}
}{
  \KOMAoptions{parskip=half}}
\makeatother
\usepackage{xcolor}
\makeatletter
\ifx\textbf\undefined\else
  \let\oldparagraph\textbf
  \renewcommand{\textbf}{
    \@ifstar
      \xxxParagraphStar
      \xxxParagraphNoStar
  }
  \newcommand{\xxxParagraphStar}[1]{\oldparagraph*{#1}\mbox{}}
  \newcommand{\xxxParagraphNoStar}[1]{\oldparagraph{#1}\mbox{}}
\fi
\ifx\textbf\undefined\else
  \let\oldsubparagraph\textbf
  \renewcommand{\textbf}{
    \@ifstar
      \xxxSubParagraphStar
      \xxxSubParagraphNoStar
  }
  \newcommand{\xxxSubParagraphStar}[1]{\oldsubparagraph*{#1}\mbox{}}
  \newcommand{\xxxSubParagraphNoStar}[1]{\oldsubparagraph{#1}\mbox{}}
\fi
\makeatother

\usepackage{longtable,booktabs,array}
\usepackage{calc}  
\usepackage{etoolbox}
\makeatletter
\patchcmd\longtable{\par}{\if@noskipsec\mbox{}\fi\par}{}{}
\makeatother

\IfFileExists{footnotehyper.sty}{\usepackage{footnotehyper}}{\usepackage{footnote}}
\makesavenoteenv{longtable}
\usepackage{graphicx}
\makeatletter
\def\maxwidth{\ifdim\Gin@nat@width>\linewidth\linewidth\else\Gin@nat@width\fi}
\def\maxheight{\ifdim\Gin@nat@height>\textheight\textheight\else\Gin@nat@height\fi}
\makeatother

\setkeys{Gin}{width=\maxwidth,height=\maxheight,keepaspectratio}

\makeatletter
\def\fps@figure{htbp}
\makeatother

\makeatletter
\@ifpackageloaded{caption}{}{\usepackage{caption}}
\AtBeginDocument{%
\ifdefined\contentsname
  \renewcommand*\contentsname{Table of contents}
\else
  \newcommand\contentsname{Table of contents}
\fi
\ifdefined\listfigurename
  \renewcommand*\listfigurename{List of Figures}
\else
  \newcommand\listfigurename{List of Figures}
\fi
\ifdefined\listtablename
  \renewcommand*\listtablename{List of Tables}
\else
  \newcommand\listtablename{List of Tables}
\fi
\ifdefined\figurename
  \renewcommand*\figurename{Figure}
\else
  \newcommand\figurename{Figure}
\fi
\ifdefined\tablename
  \renewcommand*\tablename{Table}
\else
  \newcommand\tablename{Table}
\fi
}
\@ifpackageloaded{float}{}{\usepackage{float}}
\floatstyle{ruled}
\@ifundefined{c@chapter}{\newfloat{codelisting}{h}{lop}}{\newfloat{codelisting}{h}{lop}[chapter]}
\floatname{codelisting}{Listing}

\makeatother
\makeatletter
\@ifpackageloaded{caption}{}{\usepackage{caption}}
\@ifpackageloaded{subcaption}{}{\usepackage{subcaption}}
\makeatother

\ifLuaTeX
  \usepackage{selnolig}  
\fi
\usepackage[]{natbib}

\usepackage{bookmark}

\IfFileExists{xurl.sty}{\usepackage{xurl}}{} 
\hypersetup{
  colorlinks=true,
  linkcolor={blue},
  filecolor={Maroon},
  citecolor={Blue},
  urlcolor={Blue},
  pdfcreator={LaTeX via pandoc}}

\usepackage{amsmath, amssymb, amsthm}
\usepackage{graphicx}
\usepackage{enumerate}
\usepackage{enumitem}
\usepackage{natbib}
\usepackage{url}
\usepackage{geometry}
\usepackage{booktabs} 
\usepackage{multirow} 
\usepackage{array} 
\usepackage{xcolor} 
\usepackage{colortbl}
\usepackage{makecell}
\usepackage{setspace}
\usepackage{hyperref}
\usepackage{tabularx}
\usepackage[title,toc,titletoc]{appendix}
\usepackage{bbm}
\usepackage{bm} 
\usepackage{algorithm}
\usepackage{booktabs}
\usepackage{longtable}
\usepackage{siunitx}
\usepackage{multirow}
\usepackage{algpseudocode}
\usepackage{multibib}
\usepackage{etoc}
\makeatletter
\newcommand{\appendixtableofcontents}{%
  \section*{Table of Contents}
  \@starttoc{apx}
}
\makeatother
\newcites{supp}{References of the supplementary material}

\usepackage{subcaption}
\newcommand{\anon}{1}

\newtheorem{theorem}{Theorem}
\newtheorem{assumption}{Assumption}
\newtheorem{remark}{Remark}

\newtheorem{example}{Example}

\newcommand{\R}{\mathbb{R}}
\newcommand{\E}{\mathbb{E}}

\newcommand{\myrothead}[2][75]{\rotatebox{#1}
{\makecell[c]{#2}}}

\newcommand{\Dec}{\mathsf{Dec}}

\begin{document}

\def\spacingset#1{\renewcommand{\baselinestretch}%
{#1}\small\normalsize} \spacingset{1}


\if1\anon
{
  \title{\bf ADD for Multi-Bit Image Watermarking}
  \author{An Luo and Jie Ding\\
    School of Statistics, University of Minnesota\\
luo00318@umn.edu and dingj@umn.edu}
  \date{}
  \maketitle
} \fi

\if0\anon
{
  \bigskip
  \bigskip
  \bigskip
  \begin{center}
    {\LARGE\bf 
    ADD for Multi-Bit Image Watermarking
    }
\end{center}
  \medskip
} \fi

\bigskip
\begin{abstract}
As generative models enable rapid creation of high-fidelity images, societal concerns about misinformation and authenticity have intensified. A promising remedy is multi-bit image watermarking, which embeds a multi-bit message into an image so that a verifier can later detect whether the image is generated by someone and further identify the source by decoding the embedded message. Existing approaches often fall short in capacity, resilience to common image distortions, and theoretical justification. To address these limitations,
we propose ADD (Add, Dot, Decode), a multi-bit image watermarking method with two stages: learning a watermark to be linearly combined with the multi-bit message and added to the image, and decoding through inner products between the watermarked image and the learned watermark. On the standard MS-COCO benchmark, we demonstrate that for the challenging task of 48-bit watermarking, ADD achieves 100\% decoding accuracy, with performance dropping by at most 2\% under a wide range of image distortions, substantially smaller than the 14\% average drop of state-of-the-art methods. In addition, ADD achieves substantial computational gains, with 2-fold faster embedding and 7.4-fold faster decoding than the fastest existing method. We further provide a theoretical analysis explaining why the learned watermark and the corresponding decoding rule are effective.
\end{abstract}

\noindent%
{\it Keywords:} Multi-bit Watermarking, Image Watermarking, Hypothesis Testing, Regularization
\vfill

\newpage
\spacingset{1.8}

\section{Introduction}\label{sec:intro}

In recent years, generative artificial intelligence \citep{rombachHighResolutionImageSynthesis2022,controlnet,lipman2023flow} has achieved unprecedented levels of realism and versatility, enabling rapid creation of high-fidelity images and videos~\citep{sd3,deepmind2025veo3}. At the same time, this proliferation has raised significant concerns over misinformation like DeepFake \citep{verdolivaMediaForensicsDeepFakes2020} and intellectual property infringement \citep{Sag2023Copyright,chandra2024nistSyntheticContent}. As these issues intensify, reliable methods for verifying the authenticity and provenance of digital content have become increasingly important. 

\begin{figure}[htb]
    \centering
\includegraphics[width=1.0\textwidth]{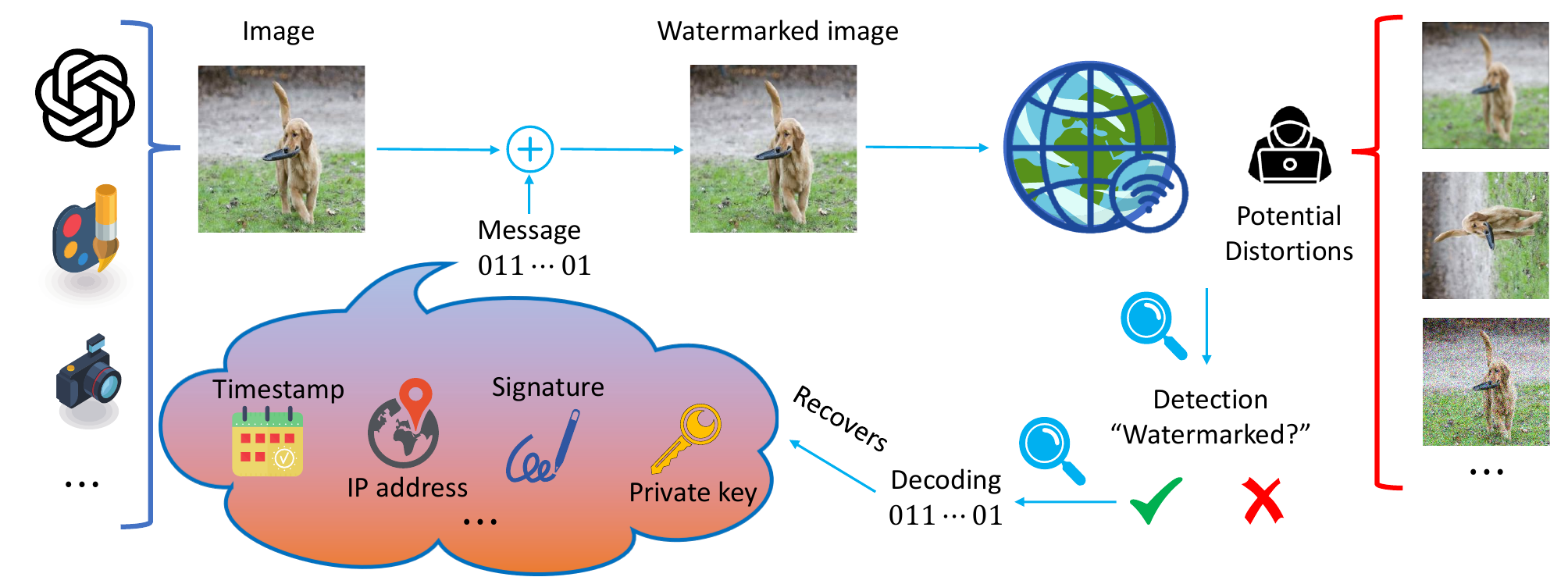}
    \caption[Overview of multi-bit watermarking]{\textbf{An overview of multi-bit image watermarking.} An image, which may be generated by AI models or created as digital artwork such as paintings or photographs, can be embedded with a multi-bit message. 
Such messages may encode information such as a timestamp, IP address, signature, or private key. 
The resulting watermarked image is then distributed over the Internet and may undergo distortions (e.g., compression or rotation). 
Given a possibly distorted image, the goal is to detect whether the image is watermarked and further decode the embedded multi-bit message to recover the information.}
    \label{fig:overview_watermark}
\end{figure}

In this context, image watermarking has emerged as a promising approach to invisibly embed information into images for purposes such as content verification and copyright protection  \citep{coxDigitalWatermarkingSteganography2007}. 
Among watermarking techniques, multi-bit watermarking (see Figure~\ref{fig:overview_watermark} for an overview) is particularly important because it enables the embedded watermark to carry information for origin tracking and attribution. 
In contrast to single-bit watermarking, which only indicates whether an image is watermarked, multi-bit watermarking embeds a message that can represent richer information, such as an owner identifier, a user fingerprint, a timestamp, or an IP address. However, increasing the message capacity also makes the problem more challenging: the watermark must remain invisible while still enabling resilient recovery of embedded message under common image distortions.

Despite substantial progress, existing multi-bit image watermarking methods still exhibit limitations. Traditional watermarking techniques embed watermark by modifying pixel values \citep{Schyndel1994ADW,NIKOLAIDIS1998385,chenQuantizationIndexModulation2001,altunOptimalSpreadSpectrum2009,jieNewPublicWatermarking2009} or adding signals in frequency domain \citep{coxSecureSpreadSpectrum1996,frequency1997,hernandezDCTdomainWatermarkingTechniques2000, al-hajCombinedDWTDCTDigital2007, aaDWTDCTSVDBasedWatermarking2008} of images. Despite their simple implementation and historical significance, these classical methods either consider only single-bit watermark, or struggle to handle intensive image distortions, such as compression and rotation \citep{balleVariationalImageCompression2018}.
 With the advent of deep learning \citep{Goodfellow-et-al-2016}, a new wave of watermarking schemes has emerged that harness the power of neural networks with encoder–decoder architectures  \citep{zhuHiddenHidingData2018, tancikStegastampInvisibleHyperlinks2020,fernandezWatermarkingImagesSelfsupervised2022,xian2024raw,sander2025watermark}, where encoders hide messages within images and decoders are designed to resiliently decode the hidden messages. These approaches can achieve better performance than traditional methods, but ensuring consistent resilience to diverse distortions still remains challenging \citep{dlwsurvey,anWAVESBenchmarkingRobustness2024}. In addition, most of them operate at low capacity (no more than 32 bits) and provide limited theoretical guidance. Some recent works have proposed watermarking methods designed specifically for certain image generative models, including diffusion models \citep{fernandezStableSignatureRooting2023a,wenTreeRingWatermarksFingerprints2023,yangGaussianShadingProvable2024,gunnUndetectableWatermarkGenerative2024} and  autoregressive models \citep{jovanovic2025watermarking}, which can achieve better resilience in some cases but are inherently tied to a particular architecture and do not work for other model families or post-hoc settings. From a theoretical perspective, watermarking has been studied mainly in single-bit settings, particularly through information-theoretic perspectives \citep{willems2000,moulin2000,MOULIN20011121,moulin2003,LiuandMoulin2003b,Liuandmoulin2003a,sion2004}, and more recently through statistical frameworks for LLM-generated text \citep{detecthumanedit,statframework,watermarkmaxcoupling}.

To address these limitations, we propose \emph{Add, Dot, Decode} (ADD), a simple yet effective multi-bit image watermarking method together with theoretical analysis that explains why it works. 
ADD embeds a multi-bit message by \emph{adding} to the image a linear combination of a learned watermark weighted by the multi-bit message; through \emph{dot products} between the watermarked image and the learned watermark, ADD \emph{decodes} the embedded message. Such watermark is learned through a training objective designed to jointly pursue image quality, decoding performance, and resilience to distortions.

The key insight underlying ADD is that the learned watermark possesses 
a specific geometric structure. Under the assumption that the image data concentrate around a low-dimensional subspace with Gaussian noise perturbation, the training objective 
learns a watermark that is orthogonal to the low-dimensional image subspace, 
with mutually orthogonal watermark components corresponding to the bits of the message. These properties ensure that the inner products between images and the watermark 
are well separated for watermarked and unwatermarked images, enabling reliable detection. 
Moreover, because each message bit is embedded with a distinct watermark component, 
the sign of the inner product between the watermarked image and that component 
reveals the bit value, enabling accurate decoding.

Building on this geometric structure, we derive detection and 
decoding rules using the likelihood principle and establish corresponding 
performance guarantees. We further show that this geometric property also holds asymptotically for a watermark learned from the corresponding finite-sample objective, and the detection and decoding performance converge to their population counterparts. 
Empirically, ADD achieves state-of-the-art performance under a wide range of image distortions, while maintaining high image quality and offering 
substantially faster embedding and decoding than competing approaches.

The outline of the paper is given as follows.
In Section~\ref{sec:formulation}, we state the problem formulation of watermarking. In Section~\ref{sec:methods}, we introduce ADD. In Section~\ref{sec:theory}, we provide a theoretical analysis on how ADD can work.
We present experiments in Section~\ref{sec:experiments} and conclude the paper in Section~\ref{sec:conclusion}. The supplementary material includes proofs and details of discussions and experiments.

\vspace{-10pt}
\section{Problem Formulation of Watermarking}\label{sec:formulation}

Consider a signal $\bm x \in \mathcal X$ and a message $\bm m \in \mathcal M$. We define a watermarking mechanism $W$ as a map that produces a watermarked signal $\tilde{\bm x} = W(\bm x, \bm m)$, where $\tilde{\bm x}$ should remain close to $\bm x$ in quality. After $\tilde{\bm x}$ being distributed through a channel, a potentially distorted version $\tilde{\bm x}'$ is observed by a verifier. On $\tilde{\bm x}'$, the verifier performs: 1) \textbf{watermark detection}, to decide whether $\tilde{\bm x}'$ is watermarked by $W$, and further 2) \textbf{watermark decoding}, to recover the message $\bm m$ embedded in $\tilde{\bm x}$ by $W$. In this paper, $\mathcal X$ is the space of images and $\bm m \in \mathcal M =\{-1,1\}^K$ is a $K$-bit message that serves as an identifier, where $K$ is a positive integer. Throughout the paper, we use $k=1,\ldots,K$ and $k\in[K]$ interchangeably. We consider the setting where an image is either unwatermarked or watermarked by $W$. 

The above formulation is generic, and we explain below with some concrete scenarios depending on who chooses 
$\bm m$, who knows 
$\bm m$, and how the message $\bm m$ can relate to identifiers.

\begin{example}[Watermark embedded by a model provider]\label{exp:modelprovider}
 A model provider Alice (e.g., OpenAI) embeds watermark in images generated for each user. A user Bob requests image generation through Alice's application programming interface (API), and Bob will always receive a watermarked image $\tilde{\bm x} = W(\bm x, \bm m)$ with an assigned identifier $\bm m$. Such a message $\bm m$ can be deterministically derived from Bob's metadata (e.g., IP address and timestamp), and assigned by Alice, while remaining unknown to Bob.  A verifier (Alice or a third-party auditor) later detects whether an image is watermarked by $W$ in order to determine whether the image is generated by Alice's model, and further decodes the embedded $\bm m$ to identify which user of Alice's model generated that image.    
\end{example}

\begin{example}[Watermark embedded by an artist]\label{exp:contentcreator}
 An artist Bob watermarks their own images, e.g., photographs or digital paintings, with their identifier 
$\bm m$, e.g., a signature or a timestamp, before releasing the images to the public. In this setting, Bob knows $\bm m$ and can act as a verifier that detect whether or not an image is distributed by them to claim ownership.   
\end{example}

\begin{example}[Watermark embedded by multiple entities]\label{exp:multorg}
A single watermarking mechanism 
$W$ can be adopted by multiple entities. To avoid collisions, e.g., same message $\bm m$ being associated with different users, the message space 
$\mathcal M$ can be partitioned so that each entity is assigned a disjoint subset. This ensures that the recovered message can be uniquely attributed to the issuing entity. 
\end{example}

When released to the public, the watermarked image $\tilde{\bm x}$ may undergo distortions, resulting in $\tilde{\bm x}' \in \mathcal X$. Such distortion includes natural noises such as compression and intentional attack such as cropping. Let $\mathcal{A}$ denote a discrete distribution over a finite set of image distortion operators $A: \mathcal{X} \rightarrow \mathcal{X}$. Given a watermarked image $\tilde{\boldsymbol{x}}$, the distorted version is $
\tilde{\boldsymbol{x}}^{\prime}=A(\tilde{\boldsymbol{x}})$ for some $A \sim \mathcal{A} .$

A watermarking mechanism $W$ should be designed for the following
objectives: 1)\textbf{Quality}: The quality of watermarked images $\tilde{\bm x}$ should remain visually close to that of unwatermarked images $\bm x$; 2)\textbf{Identifiability}: Watermarked and unwatermarked images should be accurately distinguished, and the embedded message $\bm m$ should be accurately recovered; 3)\textbf{Resilience}: The watermark should be resilient to common image
distortions.

Formally, consider the space of images $\mathcal{X}=\mathbb R^D,$
where $D$ is a positive integer. Let $\boldsymbol{x} \in \mathcal{X}$ denote an original image and let $\boldsymbol{m}= \left(m_1, \ldots, m_K\right)^{\top} \in \mathcal{M} = \{-1,1\}^K$ denote a $K$-bit watermark message. A ($K$-bit) watermarking mechanism 
$W: \mathcal{X} \times\{-1,1\}^K \rightarrow \mathcal{X}$
produces a watermarked image $\tilde{\boldsymbol{x}}=W(\boldsymbol{x}, \boldsymbol{m})$.

Watermark detection for $W$ can be formulated as a hypothesis testing problem for an image $\boldsymbol{x} \in \mathcal{X}$ to be verified:

{
\vspace{-32pt}

\begin{equation}\label{eq:hptext}
 H_0: \bm x \text{ is not watermarked} 
\quad H_1: \bm x \text{ is watermarked}  
\end{equation}

\vspace{-20pt}
}

To formalize the hypothesis problem~\eqref{eq:hptext}, let $ P_0$ denote the distribution of unwatermarked image on $\mathcal X$. For each fixed message $\bm m\in\{\pm1\}^K$, define the embedding map
$W_{\bm m}(\bm x) := W(\bm x,\bm m),$
and define the induced  distribution of watermarked images with watermark message $\bm m$ by
$P_{W\mid \bm m} :=P_0 \circ W_{\bm m}^{-1}$, i.e., 
$P_{W\mid \bm m}(E)=\mathbb P\big(W(\bm x,\bm m)\in E\big)
$ for all measurable $E\subseteq\mathcal X.$
Then the watermark detection problem can be posed as the composite hypothesis test:

{
\vspace{-32pt}

\begin{equation}\label{eq:hpdetect}
H_0:\ \bm x \sim P_0
\quad
H_1:\ \bm x \sim P_{W\mid \bm m}\ \text{for some } \bm m\in\{\pm1\}^K.
\end{equation}

\vspace{-20pt}
}

As explained in Examples~\ref{exp:modelprovider}-\ref{exp:multorg}, a verifier may have access to a dictionary $\mathcal D$ of messages embedded. This makes the detection problem different from \eqref{eq:hpdetect} as the search space of messages are potentially smaller, and it is another detection problem to be considered

{
\vspace{-32pt}

\begin{equation}\label{eq:hpdetectD}
H_0:\ \bm x \sim P_0
\quad
H_{1'}:\ \bm x \sim P_{W\mid \bm m}\ \text{for some } \bm m\in\mathcal D.
\end{equation}

\vspace{-20pt}
}

Watermark decoding for $W$ aims to build a decoder \(\Dec: \mathcal{X}\to \{-1,1\}^K\) that recovers the watermark message from the watermarked image with
$\widehat{\bm m} = \Dec (\tilde{\bm{x}}),$
with the ultimate goal of maximizing the \emph{expected bit accuracy} $\E \left[\frac{1}{K}\sum_{k=1}^K \mathbbm{1}\{\widehat m_{k}=m_{k}\}\right]$. We consider bit accuracy rather than the probability of perfect decoding for two reasons. 
First, bit accuracy is the standard metric in multi-bit watermarking and allows direct comparison with prior work. 
Second, perfect decoding is an all-or-nothing criterion that becomes increasingly stringent as the message length $K$ grows, since an error on a single bit makes the entire message incorrect, so bit accuracy provides a more informative measure of decoding performance.

\vspace{-10pt}
\section{Add, Dot, Decode (ADD)}\label{sec:methods}

\begin{figure}[htb]
    \centering
\includegraphics[width=1.0\textwidth]{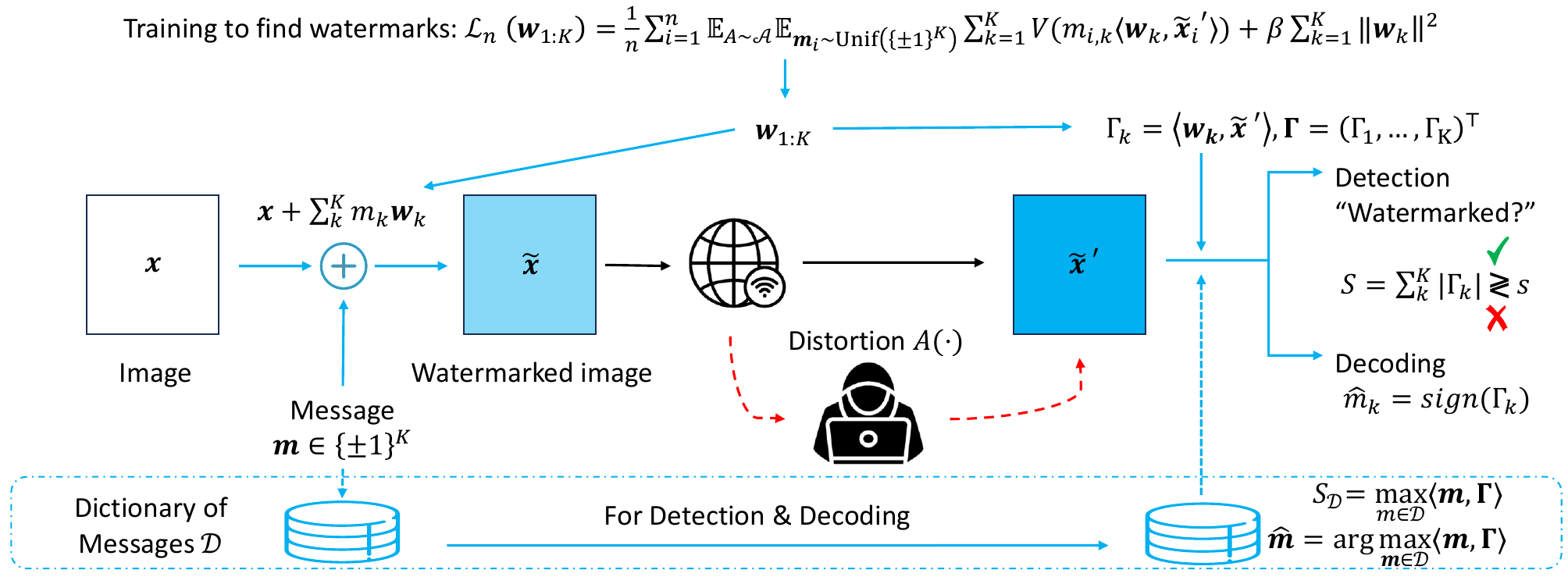}
    \caption[Overview of multi-bit watermarking]{\textbf{Overview of ADD for multi-bit image watermarking.} Given an image $\bm x\in\mathcal X$, a $K$-bit message $\bm m\in\{\pm1\}^K$ is embedded by $\tilde{\bm x}=\bm x+\sum_{k=1}^K m_{k}\bm w_k$. After distribution, the watermarked image may be distorted by an distortion operator $A$, yielding the observed image $\tilde{\bm x}'_i = A(\tilde{\bm x})$. \emph{Detection} is performed to decide whether $\tilde{\bm x}'$ is watermarked by ADD and if detected, \emph{decoding} is performed to recover $\bm m$.  Specifically, detection and decoding are based on the inner products $\Gamma_k=\left\langle\boldsymbol{w}_k, \tilde{\boldsymbol{x}}^{\prime}\right\rangle$, collected as $I=\left(\Gamma_1, \ldots, \Gamma_K\right)^{\top}$: detect with $S=\sum_{k=1}^K\left|\Gamma_k\right|$, and decode with $\hat{m}_k=\operatorname{sign}\left(\Gamma_k\right)$. As indicated in the dashed region, when an optional dictionary of embedded messages $\mathcal D\subset\{\pm1\}^K$ is available, detection and decoding can be improved with $S_{\mathcal{D}}=\max _{\boldsymbol{m} \in \mathcal{D}}\langle\boldsymbol{m}, I\rangle$ and $\hat{\boldsymbol{m}}= \arg \max _{\boldsymbol{m} \in \mathcal{D}}\langle\boldsymbol{m}, I\rangle$. The watermark $\bm w_{1:K}$ is trained from the top objective $\mathcal L_n$ as proposed in~\eqref{eq:loss1}.}
    \label{fig:overview}
\end{figure}

In this section, we propose ADD, with an overview provided in Section~\ref{sec:method_overview} (see also Figure~\ref{fig:overview}), followed by details on watermark training in Section~\ref{sec:formulation_train} and deployment (watermark embedding, decoding and detection procedures) in Section~\ref{sec:formulation_deploy}.

\vspace{-10pt}
\subsection{Overview of ADD}\label{sec:method_overview}
Our goal is to develop a watermarking mechanism that embeds a $K$-bit message $\bm m\in\{\pm 1\}^K$ into an image $\bm x$, and ensures quality, identifiability, and resilience, as discussed in Section~\ref{sec:formulation}. 

We first learn a watermark $\bm w_{1:K}$ to be added to $\bm x$ by optimizing a training objective that balances the above three goals. Then, our watermarking mechanism $W$ is additive to original image $\bm x$, i.e., $\tilde{\bm{x}} = W(\bm x, \bm m)=\bm{x} +\sum_{k=1}^{K} m_k \bm{w}_k$. We store watermark $\bm w_{1:K}$ for later detection and decoding. When an image $\bm x$ is received for detection or decoding, we compute inner products $\Gamma_k = \langle\bm w_k, \bm x \rangle, k = 1,\ldots, K$, between the image and our saved watermark $\bm w_{1:K}$. For detection, we aggregate these inner products into a test statistics $S = \sum^K_{k=1} |\Gamma_k|$ and reject $H_0$ (in \eqref{eq:hpdetect}) if $S>s$ for a threshold $s$. For decoding, we recover each bit via $\widehat{m}_k = \text{sign}(\Gamma_k), k = 1,\ldots, K$. If a message dictionary $\mathcal D$ is available, we detect with statistics $S_\mathcal D = \max_{\bm m\in\mathcal D}\ \sum_{k=1}^K m_k \Gamma_k$ and decode with $\arg\max_{\bm m\in\mathcal D}\ \sum_{k=1}^K m_k \Gamma_k$. We will explain these in detail in the rest of this section.

\vspace{-10pt}
\subsection{Training to find the desirable watermark}\label{sec:formulation_train}

To train the desirable $\bm w_{1:K}$ that satisfy the requirements on quality, identifiability, and resilience, we construct a training objective that jointly pursues these three goals. Below we explain how we incorporate each goal into our training objective. 

For quality, we include $\beta \sum_{k=1}^K\|\bm{w}_k\|^2$ with a hyperparameter $\beta > 0$, where $\|\cdot\|$ denotes the Euclidean ($\ell_2$) norm, so that we constrain the magnitude of $\bm w_{1:K}$ in terms of the squared $\ell_2$ norms. This is a common practice to penalize the magnitude of the watermark~\citep{zhuHiddenHidingData2018,tancikStegastampInvisibleHyperlinks2020, fernandezWatermarkingImagesSelfsupervised2022}, which is also a classical way to penalize the magnitude of parameters as introduced by~\cite{Hoerl01021970} and~\cite{Elasticnet}. 

For identifiability, we first explain our approach with the simplest case $K=1$. When $K=1$, the watermark decoding problem boils down to a binary classification problem, i.e., predicting the message $\bm m = m_1 =: m$ embedded in a watermarked image $\tilde{\bm x}$ with label $m \in \{\pm 1\}$. A standard solution to such binary classification problem is to train with a margin-based loss~\citep{LIN200473} of the form $V(mf(\cdot))$, where $V$ is a margin-based loss such as hinge loss, $m$ is the prediction target (here, the message embedded and to be recovered), and $f(\cdot)$ is the classification function with classification rule $\text{sign}(f)$ (here, $f(\cdot)$  should be some quantity determined by the watermarked image $\tilde{\bm x}$ and the watermark $\bm w:=\bm w_1$), i.e., predict $\widehat{m} = 1$ if $\text{sign}(f) \geq 0$ and $\widehat{m} = -1$ if $\text{sign}(f) < 0$. As described earlier in Section~\ref{sec:method_overview}, $f(\cdot)$ would be the inner product $\langle \bm w, \tilde{\bm x} \rangle$. Therefore, at $K=1$ the term for the loss that enforces identifiabiliy would be $V(m \langle \bm w, \tilde{\bm x} \rangle)$. One key difference between our watermarking at $K=1$ and binary classification is that, we take full control on the prediction target $m$, i.e., as training data we only need to independently sample $m \sim \mathrm{Unif}(\pm1)$, take them as prediction targets, and embed them through our watermarking mechanism. This requires a expectation form in the loss, i.e., $\E_{m\sim\mathrm{Unif}(\pm1)}V(m \langle \bm w, \tilde{\bm x} \rangle)$.  Extending to the case $K>1$, since decoding is going to be performed with $\text{sign}(\langle\bm w_k, \bm x \rangle)$ for each $k$, we only need to do binary classification separately for each bit $m_k$, and this leads to the term $\E_{\bm m\sim\mathrm{Unif}({\{\pm1\}}^K)}\sum^K_{k=1}V(m_k \langle \bm w_k, \tilde{\bm x} \rangle)$. Watermark detection is given by $S = \sum^K_{k=1} |\langle\bm w_k, \bm x \rangle| $ would be large for watermarked images and small for unwatermarked images, as will be explained in Section~\ref{sec:detection_decoding_dictionary}.

For resilience, we add distortion simulation in training, to ensure the decoding mechanism also works well for watermarked images under common image distortions (examples of such image distortions are provided in Section~E of the supplementary material). Specifically, we simulate the distorted watermarked image
$\tilde{\bm{x}}'$ by applying a randomly sampled distortion operator  $A\sim\mathcal{A}$ to the
watermarked image $\tilde{\bm{x}}$, and set
$\tilde{\bm{x}}' =A(\tilde{\bm{x}}).$ This yields the term $\E_{A\sim\mathcal A}\E_{\bm m\sim\mathrm{Unif}({\pm1}^K)}\sum^K_{k=1}V(m_k \langle \bm w_k, \tilde{\bm x}' \rangle)$.

As discussed above, to ensure the quality, identifiability, and resilience of the watermark $\bm w_{1:K}$, we propose our population training objective here:
\begin{equation}\label{eq:loss0}
 \mathcal{L}(\bm{w}_{1:K}) = \E_{\bm x\sim P_0} \E_{A\sim\mathcal{A}}\mathbb{E}_{\bm m  \sim \mathrm{Unif}({\{\pm1\}}^K)}\sum^{K}_{k=1}V\left(m_{k} \langle  \bm{w}_k,\tilde{\bm{x}}'\rangle\right) + \beta\sum_{k=1}^K\|\bm{w}_k\|^2,  
\end{equation}
where $P_0$ denotes the distribution of unwatermarked images $\boldsymbol{x} \in \mathcal{X}$, $\mathcal{A}$ is a discrete distribution over a finite set of image distortion operators $A: \mathcal{X} \rightarrow \mathcal{X}$, $\boldsymbol{m}=\left(m_1, \ldots, m_K\right)^{\top}$ is randomly sampled from $\operatorname{Unif}\left(\{ \pm 1\}^K\right)$,  $\tilde{\boldsymbol{x}}^{\prime}=A(\tilde{\boldsymbol{x}})=A\circ W(\boldsymbol{x}, \boldsymbol{m})$ is the distorted version of the watermarked image, and $\beta>0$ is the regularization parameter.

The finite-sample objective is obtained by replacing the expectation $\E_{\bm x\sim P_0}$ in \eqref{eq:loss0} with an average over a finite training set $\{\bm x_i\}^n_{i = 1}$ of unwatermarked images:
\begin{equation}\label{eq:loss1}
 \mathcal{L}_n(\bm{w}_{1:K}) = \frac{1}{n}\sum_{i=1}^n \E_{A\sim\mathcal{A}}\mathbb{E}_{\bm m_{i}  \sim \mathrm{Unif}({\{\pm1\}}^K)}\sum^{K}_{k=1}V\left(m_{i,k} \langle  \bm{w}_k,\tilde{\bm{x}}'_i\rangle\right) + \beta\sum_{k=1}^K\|\bm{w}_k\|^2,  
\end{equation}
where $\bm m_i = (m_{i,1},\ldots,m_{i,K})^{\top} \sim \text{Unif}\left({\{\pm 1\}^K}\right)$, and for each given $\boldsymbol{m}_i$ and $A$, the watermarked image is $\tilde{\boldsymbol{x}}_i= W(\boldsymbol{x}_i, \boldsymbol{m}_i)$ and its distorted version is $\tilde{\boldsymbol{x}}^{\prime}_i=A(\tilde{\boldsymbol{x}}_i)=A \circ W(\boldsymbol{x}_i, \boldsymbol{m}_i) .$

The finite-sample objective \eqref{eq:loss1} describes the exact optimization problem. When implementing it, we use Monte Carlo samples of the message $\bm m_i$ and distortion operator $A$ at each gradient update of stochastic gradient descent (SGD). Moreover, rather than optimizing $\bm w_{1:K}$ directly in the pixel space of high dimension, we learn it through a low-dimensional parameterization. We present the implementation of our training algorithm with pseudo code in Algorithm~\ref{alg:train} and also explain it below.

To construct the watermark $\bm w_{1:K}$, instead of directly optimizing in the high-dimensional image space, we optimize over a space with much lower dimension. We first extract features from the original image using a pretrained and frozen feature extractor $\psi: \mathcal{X} \to \mathbb{R}^{d_f}$ with output dimension \(d_f<D\). To match the watermark dimension, for each $k$ we train a watermark map $g_k: \mathbb{R}^{d_f} \to \mathbb{R}^{D}$, yielding a per-image watermark
$\bm{w}_k= g_k\bigl(\psi(\bm{x})\bigr)$. When computing the penalty term, we normalize by the dimension $D$ to ensure numerical stability. 
After training converges, we freeze the learned watermark maps ${g_{k}}$ and compute their dataset-level averages by evaluating them on all training images and averaging the resulting outputs. These averages directly construct a fixed watermark ${\bm w}_{1:K}$, which is used for deployment.

\begin{algorithm}[htb]
\caption{\label{alg:train}
Training (learn watermark for $K$ bits)}
\begin{algorithmic}[1]
\Require Training set $\{\bm x_i\}_{i=1}^n$, number of bits $K$,
feature extractor $\psi:\mathcal X\to\mathbb R^{d_f}$ (frozen),
watermark maps $\{g_{k}:\mathbb R^{d_f}\to\mathbb R^{D}\}_{k\in[K]}$ (trainable),
margin-based loss $V$, regularization $\beta>0$,
distortion sampler $\mathcal A$.
\Ensure Trained and fixed watermark $\bm w_{1:K}$ for deployment.

  \For{each minibatch $\{\bm x_i\}_{i\in\mathcal I}$}
    \State Sample message bits $\{m_{i,k}\}_{i\in\mathcal I,\,k\in[K]}$ i.i.d.\ from $\text{Unif}(\pm1)$.
    \State Compute features $f_i\gets \psi(\bm x_i) \in \mathbb R^{d_f}$ for all $i\in\mathcal I$.
      \State Form watermark (per image):
$      \bm w_{i,k}\ \gets\ g_{k}(f_i)
    \in\mathbb R^{D}, \text{ for } k =1,\ldots,K.$
    \State Embed the $K$-bit message: $
    \tilde{\bm x}_i \ \gets\ \bm x_i + \sum_{k=1}^K m_{i,k}\,\bm w_{i,k},
    i\in\mathcal I.$
    \State Apply a distortion: Sample $A\sim\mathcal A$ and set $\tilde{\bm x}'_i \gets A(\tilde{\bm x}_i)$ for all $i\in\mathcal I$.
    \State Compute total loss on the minibatch:
    \[
    \mathcal L_{\mathrm{batch}}
    \ \gets\
    \frac{1}{|\mathcal I|}\sum_{i\in\mathcal I}\sum_{k=1}^K
    V\!\big(m_{i,k}\,\langle \bm w_{i,k},\tilde{\bm x}'_i\rangle\big)
    \;+\;
    \beta\cdot\frac{1}{|\mathcal I|}\frac{1}{D}\sum_{i\in\mathcal I}\sum_{k=1}^K \|\bm w_{i,k}\|^2.
    \]
    \State Backpropagate $\nabla \mathcal L_{\mathrm{batch}}$ through $\{g_{k}\}_{k\in[K]}$ and update parameters with SGD.
  \EndFor

\State Freeze the trained $\{g_{k}\}_{k\in[K]}$ and compute dataset-level averages:
\[
\bm w_{k}
\ \gets\
\frac{1}{n}\sum_{i=1}^n g_{k}\!\big(\psi(\bm x_i)\big),
\text{ for } k =1,\ldots,K.
\]
\State \Return $\bm w_{1:K}$.
\end{algorithmic}
\end{algorithm}

\vspace{-10pt}
\subsection{Deployment of ADD}\label{sec:formulation_deploy}
Deployment consists of watermark embedding and watermark detection and decoding.

To embed watermark (see Algorithm~\ref{alg:embed} for pseudo code), we do $\tilde{\bm{x}} = \bm{x} +\sum_{k=1}^{K} m_k \bm{w}_k,$ where \(\bm{w}_1,\dots,\bm{w}_K\) is the watermark trained in the way mentioned in Section~\ref{sec:formulation_train}.
\begin{algorithm}[htb]
\caption{\label{alg:embed}
Watermark embedding}
\begin{algorithmic}[1]
\Require Input image $\bm x$, number of bits $K$,
learned watermark $\bm w_{1:K}$ (from Algorithm~\ref{alg:train}),
message $\bm m\in\{\pm1\}^K$ (given or sampled).
\Ensure Watermarked image $\tilde{\bm x}$.
\State $\tilde{\bm x} \gets \bm x + \sum_{k=1}^K m_k\,{\bm w}_k$.

\State \Return $\tilde{\bm x}$.
\end{algorithmic}
\end{algorithm}
\vspace{-10pt}

To do watermark detection and decoding (see Algorithm~\ref{alg:detect_decode} for pseudo code), we first obtain the inner products
$\Gamma_k := \langle \bm{w}_k, \bm{x}  \rangle$, $k=1,\dots,K$. We use $S = \sum_{k=1}^K |\Gamma_k|$ as test statistics for watermark detection as defined in \eqref{eq:hpdetect}, i.e., reject $H_0$ if $S>s$, where $s$ is a threshold determined before deployment. Given a received image \(\bm x\), we decode for each bit $k$ with $\widehat{m}_k = \text{sign}\big(\Gamma_k\big)$, which is also derived in Section~\ref{sec:detection_decoding_dictionary}. If $\mathcal D$ is available, We use $S_{\mathcal D} = \max_{\bm m \in \mathcal D}\sum^K_{k=1} m_k \Gamma_k$ as test statistics for watermark detection as defined in \eqref{eq:hpdetectD} , i.e., reject $H_0$ if $S_{\mathcal D}>s_\mathcal{D}$, where $s_\mathcal{D}$ is another threshold. Then decode with $\widehat{\bm m} = \arg \max _{\bm m \in \mathcal D}\sum_{k=1}^Km_k \Gamma_k $. These test statistics and decoding rules are derived based on generalized likelihood ratio tests in Section~\ref{sec:detection_decoding_dictionary}.
\begin{algorithm}[htb]
\caption{\label{alg:detect_decode}
Watermark detection and decoding}
\begin{algorithmic}[1]
\Require Received image $\bm x$ (possibly watermarked),
number of bits $K$,
fixed watermark $\{{\bm w}_k\}_{k=1}^K$,
detection threshold $s>0$. \emph{optional} message dictionary $\mathcal D$ and threshold $s_{\mathcal D}>0$.
\Ensure Detection decision $\hat d\in\{0,1\}$,
and decoded message $\widehat{\bm m}$ (if detected).
\State $\Gamma_k \gets \langle \bar{\bm w}_k, \bm x \rangle$, for $k=1,\dots,K$.
\If{$\mathcal D$ is provided}
  \State $S_{\mathcal D} \gets \max_{\bm m\in\mathcal D}\ \sum_{k=1}^K m_k \Gamma_k$.
  \State $\hat d \gets \mathbbm{1}\{S_{\mathcal D}>s_{\mathcal D}\}$. \Comment{$\hat d=1$ means watermark detected}
    \State $\widehat{\bm m} \gets \arg\max_{\bm m\in\mathcal D}\ \sum_{k=1}^K m_k \Gamma_k$.
    \Comment{Watermark decoding}
\Else
  \State $S \gets \sum_{k=1}^K |\Gamma_k|$.
  \State $\hat d \gets \mathbbm{1}\{S>s\}$ \Comment{$\hat d=1$ means watermark detected}
      \State $\widehat{m}_k \gets \mathrm{sign}(\Gamma_k)$, for $k=1,\dots,K$.
        \Comment{Watermark decoding}
\EndIf
\State \Return $\hat d$, $\widehat{\bm m}$ (if $\hat{d} = 1$).
\end{algorithmic}
\end{algorithm}
\vspace{-10pt}

\vspace{-10pt}
\section{Theoretical Analysis}\label{sec:theory}

In this section, we provide theoretical insights into our proposed watermarking method. Specifically, in Section~\ref{sec:orth_watermarks} we elucidate how our method leverages a low-dimensional data assumption and produces watermark that is orthogonal (or nearly orthogonal) to the low-dimensional subspace of images. Based on this property of watermark, in Section~\ref{sec:detection_decoding_dictionary} we derive principled detection and decoding rules, and further analyze the performance. 

\vspace{-10pt}
\subsection{Existence and properties of the learned watermark}\label{sec:orth_watermarks}
\subsubsection{Low-dimensional image data and loss assumptions}\label{sec:prelim}
Empirical studies suggest that high-dimensional data, such as natural images, lie near a low-dimensional manifold~\citep{Goodfellow-et-al-2016,pope2021the}. Motivated by this, we adopt the following assumption for image data we consider:
\begin{assumption}\label{asp:lowdim}
Let $d< D - K$ be a positive integer and  $B\in\mathbb R^{D\times d}$ be a matrix with full column rank.
Define
$\mathcal U$ the column space of $B$, $\mathcal U^\perp$ the null space of $B^{\top}$, and $
\Pi_{\mathcal U}, \Pi_{\mathcal U^\perp}$ the projections onto $\mathcal U$ and $\mathcal U^\perp$.
The image data follow the perturbed low-dimensional model
\begin{equation}\label{eq:gauss_model_Kbit}
\bm X  = B\bm Z + \bm \epsilon,\bm Z\sim\mathcal N(0,\Sigma_{\bm Z}),\bm \epsilon \sim \mathcal{N}(0,\sigma_\epsilon^2 \mathbf{I}_D),
\end{equation}
where $\mathcal N(\cdot,\cdot)$ denotes a Gaussian distribution with the
first argument being the mean vector and the second the covariance matrix, $\bm Z$ and $\bm \epsilon$ are independent, $\Sigma_{\bm Z}$ is a positive definite $d\times d$ matrix, $\mathbf{I}_D$ is the $D\times D$ identity matrix, and $\sigma_\epsilon\geq0$.
\end{assumption}
We provide empirical evidence consistent with Assumption~\ref{asp:lowdim} in Section~J of the supplementary material, showing that the pretrained feature vectors of images concentrate near a low-dimensional linear subspace.

Below we propose the population objective and the finite-sample objective for theoretical analysis. The only difference from~\eqref{eq:loss0} and~\eqref{eq:loss1} we proposed in Section~\ref{sec:methods} is that we set $\mathcal A$ as the degenerate distribution that assigns probability $1$ to the identity operator in the training objectives for tractability.

\textbf{Population objective.}
Let $\bm m=(m_1,\dots,m_K)\in\{\pm1\}^K$ be i.i.d.\ discrete uniform over $\{\pm 1\}^K$ and independent of $\bm X$.
For watermark $\bm w_{1:K}=(\bm w_1,\dots,\bm w_K)\in(\R^D)^K$, the watermarked image is
$\tilde{\bm X}:=\bm X+\sum_{j=1}^K m_j\bm w_j.$
The population objective is given by
\begin{equation}\label{eq:L_pop_nongauss_fixed}
\mathcal L(\bm w_{1:K})
:=
\mathbb E_{\bm X,\bm m}\!\left[
\sum_{k=1}^K
V\!\left(
m_k\big\langle \bm w_k,\tilde{\bm X}\big\rangle
\right)
\right]
+\beta\sum_{k=1}^K\|\bm w_k\|^2.
\end{equation}

\textbf{Finite-sample objective.}
Given i.i.d.\ sample $\bm x_1,\dots,\bm x_n$ from $\bm X$, for each $\bm x_i$, let $\bm m_i=(m_{i,1},\dots,m_{i,K})\in\{\pm1\}^K$ be i.i.d.\ discrete uniform over $\{\pm 1\}^K$ and independent of $\bm X$. The finite-sample objective is given by
\begin{equation}\label{eq:L_emp_nongauss_fixed}
\mathcal L_n(\bm w_{1:K})
:=
\frac{1}{n}\sum_{i=1}^n
\mathbb E_{\bm m_i}\!\left[
\sum_{k=1}^K
V\!\left(
m_{i,k}\Big\langle \bm w_k,\,
\bm x_i+\sum_{j=1}^K m_{i,j}\bm w_j
\Big\rangle
\right)
\right]
+\beta\sum_{k=1}^K\|\bm w_k\|^2.
\end{equation}
By construction, $\mathcal L(\bm w_{1:K})=\E_{\bm x_{1:n}}[\mathcal L_n(\bm w_{1:K})]$.

\begin{assumption}\label{asp:marginV}
 $V: \R \to \R $ is convex, bounded below, and not affine in $\mathbb R$.  
\end{assumption}

\begin{assumption}\label{asp:marginVlip}
 $V$ is $L$-Lipschitz for some $L>0$, i.e. $|V(a)-V(b)|\le L|a-b|\text{ for all }a,b\in\R,$ or 
$\partial V(t)\subseteq[-L,0],\forall t\in\mathbb{R},$ where $\partial V(t)$ denotes the subdifferential of $V$ at $t$, i.e.
$\partial V(t)
:=
\bigl\{g\in\mathbb{R}:\;
V(s)\ge V(t)+g\,(s-t)\ \text{for all } s\in\mathbb{R}
\bigr\}.$
\end{assumption}

Here we introduce a one-dimensional population loss that will be used in the theoretical analysis.  Let $Z\sim\mathcal N(0,1)$ and define
\begin{equation}\label{eq:phi_h_pop_def}
\phi(r):=\E\,V\!\bigl(r+\sigma_\epsilon\sqrt r\,Z\bigr),
\qquad
h_{\mathrm{pop}}(r):=\phi(r)+\beta \,r,
\qquad r\ge0.
\end{equation}

\begin{assumption}\label{asp:Runique}
$h_{\text{pop}}(r)$ admits a unique  minimizer $r_\star>0$. 
\end{assumption}

\begin{remark}[When does Assumption~\ref{asp:Runique} hold?]\label{rem:Rstar_unique}
We consider two common margin-based losses:
\textbf{1) Hinge loss $V(x) = (1-x)_+$}. A sufficient condition is $\sigma^2_\epsilon < 4$ and $\beta < 1$.  \textbf{2) Logistic loss $V(x) = \log(1+e^{-x})$}. A sufficient condition is $\sigma_\epsilon^2<-4+2\sqrt6$  and 
$\beta <\frac12-\frac{\sigma_\epsilon^2}{8}$. The insight is that to ensure Assumption~\ref{asp:Runique}, $\sigma^2_\epsilon$ and $\beta$ should not be too large. A detailed discussion on these results is in Section~C of the supplementary material.
\end{remark}

\begin{assumption}\label{asp:hinge&logistic}
 Either $V$ is hinge loss with $\sigma^2_\epsilon < 4$ and $\beta < 1$, or $V$ is logistic loss with $\sigma_\epsilon^2<-4+2\sqrt6$  and 
$\beta <\frac12-\frac{\sigma_\epsilon^2}{8}$.
\end{assumption}

Remark~\ref{rem:Rstar_unique} says that Assumption~\ref{asp:hinge&logistic} implies Assumption~\ref{asp:Runique}. Because hinge loss and logistic loss satisfy Assumptions~\ref{asp:marginV}-\ref{asp:marginVlip}, Assumption~\ref{asp:hinge&logistic} also implies Assumptions~\ref{asp:marginV}-\ref{asp:marginVlip}.

\subsubsection{Population objective: watermark perfectly orthogonal to $\mathcal U$}
\label{sssec:Kbit_population}

We first establish the properties of the minimizer of the population objective~\eqref{eq:L_pop_nongauss_fixed}.

\begin{theorem}[Existence and properties of watermark learned from $\mathcal L$]
\label{thm:Kbit_pop_core}
Under Assumptions~\ref{asp:lowdim} and \ref{asp:marginV}, and assuming $\sigma_\epsilon>0$,
there exists at least one minimizer of $\mathcal L$ in $(\mathbb R^D)^K$. Let $\bm w_{1:K}^\star$ be any minimizer of $\mathcal L$. Then $\|\bm w_k^\star\|^2$ is a minimizer of $h_{\text{pop}}$, 
$\bm w_k^\star\in \mathcal U^\perp$ and $\langle \bm w_k^\star, \bm w_j^\star\rangle=0$ for all $j,k\in [K]$ with $j \neq k$. Moreover, under Assumption~\ref{asp:Runique},
$\|\bm w_1^\star\|^2=\cdots=\|\bm w_K^\star\|^2=r_\star>0$.
\end{theorem}

\begin{theorem}[Unique but trivial minimizer of $\mathcal L$]
\label{thm:Kbit_unique}
Under Assumptions~\ref{asp:marginV} and \ref{asp:marginVlip}, assuming that $\beta> K\cdot L$ and $\E\|\bm X\|<\infty$,  it follows that $\mathcal L$ admits a unique but trivial global minimizer
$\bm 0.$

\end{theorem}

\subsubsection{Finite-sample objective: watermark nearly orthogonal to $\mathcal U$}
\label{ssec:finite_emp_gauss_K}

We now establish the results for the minimizer of the finite-sample objective~\eqref{eq:L_emp_nongauss_fixed}.
Throughout the paper, we write $a \lesssim b$ if there exists a constant $c>0$ such that $a \le c\, b$; $a \gtrsim b$ if there exists a constant $c>0$ such that $a \ge c\, b$; $a\asymp b$ if $a \lesssim b$ and $a \gtrsim b$.

\begin{theorem}[Existence and properties of watermark learned from $\mathcal L_n$]
\label{thm:finite_sample_geometry}
Under Assumptions~\ref{asp:lowdim} and \ref{asp:hinge&logistic}, there exists at least one minimizer of $\mathcal L_n$ in $(\mathbb R^D)^K$. Let $\bm w^\star_{1:K,n}$ be any minimizer of $\mathcal L_n$. Then $\bm w^\star_{1:K,n}\in \mathcal W := \Big\{\bm w_{1:K} \in (\R^D)^K : \sum_{k=1}^K \|\bm w_k\|^2 \le R^2\Big\}$, where $R^2=K(V(0) - \inf V)/\beta$.
There exists a constant $\tau>0$ that does not depend on $n$, such that if $\varepsilon\leq\tau$ and  $\sup_{\bm w_{1:K} \in \mathcal W} |\mathcal L_n(\bm w_{1:K}) - \mathcal L(\bm w_{1:K})| \le \varepsilon $, the following hold:

\begin{enumerate}[label=(\roman*),leftmargin=2em]

\item \textbf{Nontriviality.} $\min_{k\in[K]} \|\bm w^\star_{k,n}\|^2 \ \ge\ \frac{r_\star}{2}.$

\item \textbf{Radius concentration.}
$\max_{k\in[K]} \big|\|\bm w^\star_{k,n}\|^2-r_\star\big|
\ \lesssim \
\sqrt{\varepsilon}.$

\item \textbf{Near-orthogonality to the image subspace $\mathcal U$.}
$\max_{k\in[K]}\|\Pi_\mathcal U\bm w^\star_{k,n}\|
\ \lesssim \
\sqrt{\varepsilon}.$

\item \textbf{Mutual near-orthogonality.} \emph{Case (A): logistic $V$ with $\sigma_\epsilon\ge 0$ or hinge $V$ with $\sigma_\epsilon>0$.} $\max_{k\neq j}|\langle \bm w^\star_{k,n},\bm w^\star_{j,n}\rangle|
\ \lesssim \
\varepsilon^{1/4}+\varepsilon^{1/2}+\varepsilon.$ \emph{Case (B): hinge $V$ with $\sigma_\epsilon=0$.} $\max_{k\neq j}|\langle \bm w^\star_{k,n},\bm w^\star_{j,n}\rangle|
\ \lesssim \
\sqrt{\varepsilon}+\varepsilon.$
\end{enumerate}
 Furthermore, for any $\delta \in (0,1)$ and $\varepsilon_n(\delta)
:=
L R\sqrt{K\mathrm{tr}(\Sigma_{\bm X})}\,\left(
\frac{4+\frac{25}{3}\sqrt{\log(\frac{4}{\delta})}}{\sqrt n}
+
\frac{75\log(2e)}{2\log 2}\frac{\log(\frac{4n}{\delta})\log(\frac{4}{\delta})}{n}
\right)$ with $\Sigma_{\bm X}:=B\Sigma_{\bm Z}B^\top+\sigma_\epsilon^2 \mathbf{I}_D$,  we have $\mathbb P\Big(\sup_{\bm w_{1:K} \in \mathcal W} |\mathcal L_n(\bm w_{1:K}) - \mathcal L(\bm w_{1:K})| \le \varepsilon_n(\delta)\Big)\geq 1-\delta$ over the training sample $\{\bm x_i\}_{i=1}^n \overset{\text{i.i.d.}}{\sim} \bm X$. Consequently, with probability at least $1-\delta$, the bounds in (i)--(iv) hold with $\varepsilon$ replaced by $\varepsilon_n(\delta)$.

\end{theorem}

\begin{remark}[Comparison to population result]
Theorem~\ref{thm:Kbit_pop_core} gives \emph{exact} properties that hold for \emph{every} population minimizer: $\bm w^\star_k \in \mathcal U^\perp$, $\langle \bm w^\star_k, \bm w^\star_j\rangle = 0$, $\|\bm w^\star_k\| = r_\star$. Theorem~\ref{thm:finite_sample_geometry} gives the corresponding statement for \emph{empirical} minimizers of $\mathcal L_n$:
with high probability over the training sample, the same geometric relations hold up to errors controlled by
$\varepsilon_n(\delta)$.
\end{remark}

Similar to the result in Theorem~\ref{thm:Kbit_unique}, one can show that if $\beta>K\cdot L$, the finite-sample objective $\mathcal L_n$ admits the unique but trivial minimizer $\bm 0$. Therefore in practice we should set $\beta<K\cdot L$.

\vspace{-10pt}
\subsection{Watermark detection and decoding: derivation and analysis}
\label{sec:detection_decoding_dictionary}

In this section, we suppose that $\bm w_{1:K}$ has been learned from the objectives~\eqref{eq:L_pop_nongauss_fixed}  or~\eqref{eq:L_emp_nongauss_fixed} and is fixed for deployment. We discuss the detection and decoding rule under the watermarking mechanism $W(\bm x,\bm m)=\bm x +\sum^K_{k= 1} m_k\bm w_k$, based on the inner products $\bm \Gamma = (\Gamma_1,\ldots,\Gamma_K)^\top$, where $\Gamma_k = \langle \bm w_k,\bm x\rangle$ for $k = 1,\ldots, K$. In Section~\ref{sec:ddpop},  we assume $\bm w_{1:K}$ is learned from the population objective $\mathcal L$ and derive appropriate detection and decoding rules based on the likelihood principle. In Section~\ref{ssec:detection_with_learned_w}, we apply these rules to the finite-sample minimizer $\bm w^\star_{1:K,n}$ and show that, as $n\to\infty$, they satisfy desirable asymptotic properties.

\subsubsection{Detection and decoding rules for the oracle watermark}\label{sec:ddpop}

Here we assume that the watermark
$\bm w_{1:K}$ is learned from the population objective $\mathcal L$
and satisfies the properties established in Theorem~\ref{thm:Kbit_pop_core}.
We refer to it as \emph{oracle watermark}, as stated in the following assumption.
\begin{assumption}\label{asp:idealw}
The watermark $\bm w_{1:K}$ satisfies $\bm w_k \in \mathcal U^\perp$, $\|\bm w_k\|^2=r_\star>0$ and $\langle \bm w_k, \bm w_j\rangle=0$ for all $j,k\in [K]$ with $j \neq k$, and $\bm w_{1:K}$ is called oracle watermark.
\end{assumption}
\begin{theorem}[Distribution of $\bm \Gamma$ under $H_0$ and the alternatives]\label{thm:ikdist}
Under Assumptions~\ref{asp:lowdim} and \ref{asp:idealw},  $\bm \Gamma \sim \mathcal N(\bm 0, \sigma^2 \mathbf{I}_K)$ under $H_0$ and $\bm \Gamma \sim \mathcal N(\mu \bm m, \sigma^2 \mathbf{I}_K)$ under $H_1$ or $H_{1'}$, where $\mu := r_\star$ and $\sigma^2 := r_\star\sigma^2_\epsilon$. When $\sigma_\epsilon^2=0$, $\bm \Gamma=0$ a.s. under $H_0$ and $\bm \Gamma=\mu \bm m$ a.s. under $H_1$ or $H_{1'}$.
\end{theorem}
With Theorem~\ref{thm:ikdist} and $\sigma^2_\epsilon>0$, the watermark detection problems~\eqref{eq:hpdetect} and~\eqref{eq:hpdetectD} can be reduced to the following ones:
For watermark detection with no information on what $\bm m$ is embedded,

{
\vspace{-32pt}

\begin{equation}\label{eq:hywithnoD_Ik}
H_0: \bm \Gamma \sim \mathcal N(\bm 0, \sigma^2 \mathbf{I}_K) 
\quad H_1: \bm \Gamma \sim \mathcal N(\mu \bm m, \sigma^2 \mathbf{I}_K) \text{ for some message $\bm m \in \{\pm 1\}^K$}.
\end{equation}

\vspace{-20pt}
}

For watermark detection with a known dictionary $\mathcal D\subset\{\pm1\}^K$ for embedded messages, 

{
\vspace{-32pt}

\begin{equation}\label{eq:hywithD_Ik}
H_0: \bm \Gamma \sim \mathcal N(\bm 0, \sigma^2 \mathbf{I}_K)  
\quad H_{1'}:\bm \Gamma \sim \mathcal N(\mu \bm m, \sigma^2 \mathbf{I}_K) \text{ for some message $\bm m \in \mathcal D$}.
\end{equation}

\vspace{-20pt}
}

With Theorem~\ref{thm:ikdist} and $\sigma^2_\epsilon=0$, the detection problem is degenerate and the bit accuracy would always be $1$.

\begin{theorem}[Test statistic and decoding rule for \eqref{eq:hywithnoD_Ik}]\label{thm:unconstr_LRT}
Under the same conditions as in Theorem~\ref{thm:ikdist} with $\sigma^2_\epsilon>0$, the generalized likelihood ratio test (GLRT) for~\eqref{eq:hywithnoD_Ik} is based on
the test statistic $S:=\sum_{k=1}^K |\Gamma_k|$ with the rejection region $\{S> s\}$ for a threshold $s$,
and the corresponding maximum likelihood estimator of $\bm m$ is
$\widehat{\bm m}^{\,\mathrm{sign}}
:=\big(\mathrm{sign}(\Gamma_1),\dots,\mathrm{sign}(\Gamma_K)\big)^{\top}$, which is the decoded message. 
\end{theorem}

\begin{theorem}[Test statistic and decoding rule for \eqref{eq:hywithD_Ik}]\label{thm:dict_LRT}
Fix a dictionary $\mathcal D\subset\{\pm1\}^K$ of size $|\mathcal D|\in\{1,\dots,2^K\}$. Under the same conditions as in Theorem~\ref{thm:ikdist} with $\sigma^2_\epsilon>0$, the GLRT for~\eqref{eq:hywithD_Ik} is based on
the test statistic $S_{\mathcal D}:=\max_{\bm m\in\mathcal D}\ \langle \bm m,\bm \Gamma\rangle
=\max_{\bm m\in\mathcal D}\ \sum_{k=1}^K m_k \Gamma_k$ with the rejection region $\{S_{\mathcal D}> s_{\mathcal D}\}$ for a threshold $s_{\mathcal D}$,
and the corresponding maximum likelihood estimator of $\bm m$ is
$\widehat{\bm m}^{\mathcal D}:=\arg\max_{\bm m\in\mathcal D}\ \langle \bm m,\bm \Gamma\rangle$, which is the decoded message.
\end{theorem}

For any message
$\bm m\in\{\pm1\}^K$ denote the bit accuracy of a decoder $\widehat{\bm m} = (\widehat{m}_1,\ldots,\widehat{m}_K)^{\top}$ by $\textsc{ba}(\widehat{\bm m},\bm m):=\frac{1}{K}\sum_{k=1}^K
\mathbbm 1\{\widehat{m}_k=m_k\}$.
For $\bm m,\bm m'\in\{\pm1\}^K$, denote the Hamming distance $d_H(\bm m,\bm m'):=\sum_{k=1}^K \mathbbm 1\{ m_k \neq m_k' \}$ and $
d_{\min}:=\min_{\bm m\neq \bm m'\in\mathcal D} d_H(\bm m,\bm m').$

\begin{theorem}[Bit accuracy and a sufficient condition for improvement from $\mathcal D$]
\label{thm:dict_vs_sign_BA} Under the same conditions as in Theorem~\ref{thm:ikdist} with $\sigma_\epsilon>0$,
for any $\bm m \in \{\pm1\}^K$, $\E\!\left[\textsc{ba}(\widehat{\bm m}^{\,\mathrm{sign}},\bm m)\mid H_1,\bm m\right]=\Phi(\frac{\mu}{\sigma})$, where $\Phi(\cdot)$ denotes the cdf of a standard normal random variable, and for any  $\bm m \in\mathcal D$,
$\E\!\left[\textsc{ba}(\widehat{\bm m}^{\mathcal D},\bm m)\mid H_{1'},\bm m\right]
\ \ge\ 1-(|\mathcal D|-1)\,\Phi\!\big(-\frac{\mu}{\sigma}\sqrt{d_{\min}}\big).$
Furthermore, a sufficient condition for $\E\!\left[\textsc{ba}(\widehat{\bm m}^{\mathcal D},\bm m)\mid H_{1'},\bm m\right]
\ge
\E\!\left[\textsc{ba}(\widehat{\bm m}^{\,\mathrm{sign}},\bm m)\mid H_{1},\bm m\right]$ uniformly over $\bm m\in\mathcal D$ is
$|\mathcal D|
\ \le\
\frac{\mu^2\sqrt{d_{\min}}}{\mu^2+\sigma^2}\,
\exp\!\Big(\frac{\mu^2}{2\sigma^2}(d_{\min}-1)\Big)$. 
\end{theorem}

\textbf{Practical implication for $|\mathcal D|$ and $d_{\min}$.}  
Since $1 \le d_{\min} \le K$, the sufficient condition in Theorem~\ref{thm:dict_vs_sign_BA} implies that
$|\mathcal D| \;\lesssim\; \exp\!\left(\frac{(\mu/\sigma)^2}{2}\, d_{\min}\right).$
Thus the admissible size of the dictionary can grow exponentially with $d_{\min}$. 
In particular, if $d_{\min}\asymp K$, then
$|\mathcal D| \;\lesssim\; \exp\!\left(cK\right)$
for some constant $c>0$, meaning that $|\mathcal D|$ can grow exponentially in $K$ while still ensuring higher expected bit accuracy than decoding without $\mathcal D$.

\textbf{Practical implication for $\beta$.} There is a trade-off from $\beta$ between identifiability (bit accuracy, the larger the better, given by $\Phi(\frac{\mu}{\sigma})=\Phi(\frac{\sqrt{r_\star}}{\sigma_\epsilon})$) and quality ($\|\tilde{\bm x} -\bm x\|^2$, the smaller the better, given by $\|\tilde{\bm x} -\bm x\|^2 = \|\sum m_k\bm w^\star_k\|^2 = Kr_\star$): since $r_\star$ decreases with $\beta$ (as indicated in  Section~C.2 of the supplementary material), a larger $\beta$ will result in worse identifiability and better quality for the learned watermark.

To compare the two test statistics $S=\sum_{k=1}^K |\Gamma_k|$ and
$S_{\mathcal D}=\max_{\bm m\in\mathcal D}\langle \bm m,\bm \Gamma\rangle$, a natural way is to evaluate their true positive rates (TPRs) at a common false positive rate (FPR)~$\alpha$, i.e., to compare
$\mathbb P(S> s_\alpha^\star\mid H_1)$ and $\mathbb P(S_{\mathcal D}> t_\alpha^\star\mid H_{1'})$ where
$s_\alpha^\star$ and $t_\alpha^\star$ are the exact $(1-\alpha)$-quantiles of $S$ and $S_{\mathcal D}$ under $H_0$. Here $H_{1'}$ restricts the message to $\mathcal D\subset{\pm1}^K$, so this comparison is valid only when the true message is in $\mathcal D$. 
While these quantiles are well-defined under the hypothesis testing problems~\eqref{eq:hywithnoD_Ik} and~\eqref{eq:hywithD_Ik}, neither admits a simple closed form:
$S$ is a sum of folded-normal variables, and $S_{\mathcal D}$ is the maximum of a generally correlated Gaussian family indexed by $\mathcal D$ (with correlation determined by the geometry of $\mathcal D$).
As a result, an exact equal-FPR comparison is not available in closed form. Instead, we can compare the TPR or type~II error under conservative FPRs, as discussed in Section~D of the supplementary material. In practice, for a target FPR $\alpha$, the thresholds $s_\alpha^\star$ and $t_\alpha^\star$ can be set empirically as $(1-\alpha)$-quantiles of $S$ and $S_{\mathcal D}$ computed on a held-out calibration set of unwatermarked images.

\subsubsection{Detection and decoding with finite-sample watermark}
\label{ssec:detection_with_learned_w}

Watermark detection is performed on a \emph{single} test image $\bm x$ using the inner products
$\Gamma_k=\langle \bm w_k,\bm x\rangle$ and the statistic $S=\sum_{k=1}^K|\Gamma_k|$
(Section~\ref{sec:formulation_deploy}).
Consequently, the usual ``$n\to\infty$'' asymptotics in hypothesis testing does not refer to the number of test
samples. Instead, the natural $n$ for asymptotic here is the number of \emph{training} images in the finite-sample objective $\mathcal L_n$. Let $\bm w^\star_{1:K,n}=(\bm w^\star_{1,n},\dots,\bm w^\star_{K,n})^{\top}$ be a minimizer of the finite-sample objective
$\mathcal L_n$ in~\eqref{eq:L_emp_nongauss_fixed}. We emphasize here that the test image $\bm x$ is independent of the training data $\{\bm x_i\}^n_{i=1}$.

Specifically, we directly replace $\bm w_{1:K}$ in $\bm \Gamma$ and $S$ with $\bm w^\star_{1:K,n}$, and consider inner products $\bm \Gamma_n : = (\Gamma_{1,n},\ldots,\Gamma_{K,n})^{\top}$, where $\Gamma_{k,n}:=\langle \bm w^\star_{k,n}, \bm x\rangle$, and detection statistic $S_n:=\sum_{k=1}^K |\Gamma_{k,n}|.$
In this section we characterize the watermark detection  and decoding problem given the finite-sample minimizer $\bm w^\star_{1:K,n}$ and show that,
as $n\to\infty$, the FPR and TPR of the test based on $S_n$ converge to those of the oracle detection problem~\eqref{eq:hywithnoD_Ik}, and the bit accuracy of the corresponding decoder converges to its oracle counterpart.

\begin{theorem}[Distribution of $\bm \Gamma_n$ under $H_0$ and the alternatives]\label{thm:ikdist_n}
Under Assumption~\ref{asp:lowdim}, $\bm \Gamma_n \sim \mathcal N(\bm 0,\Sigma_{n})$ under $H_0$ and $\bm \Gamma_n \sim \mathcal N(G_n\bm m,\Sigma_{n})$ under $H_1$ or $H_{1'}$, where $G_n$ is the $K\times K$ matrix whose $(k,j)$-th entry is
$\langle \bm w^\star_{k,n}, \bm w^\star_{j,n} \rangle$, and
$\Sigma_{n}$ is the $K\times K$ matrix whose $(k,j)$-th entry is
${\bm w^{\star}_{k,n}}^\top \Sigma_{\bm X} \bm w^{\star}_{j,n}$.
\end{theorem}

With Theorem~\ref{thm:ikdist_n}, the watermark detection problem~\eqref{eq:hpdetect} can be reduced to the following one:{
\vspace{-20pt}
\begin{equation}\label{eq:hywithnoD_Ik_n}
H_0: \bm \Gamma_n \sim \mathcal N(\bm 0,\Sigma_{n})
\quad H_1: \bm \Gamma_n \sim \mathcal N(G_n\bm m,\Sigma_{n}) \text{ for some message $\bm m \in \{\pm 1\}^K$}.
\end{equation}

\vspace{-32pt}}

Following the detection and decoding rules derived in Theorem~\ref{thm:unconstr_LRT}, reject $H_0$ when $S_n>s$ for a threshold $s$, and decode with $\widehat{\bm m}^{\mathrm{sign}}_{n}: = (\mathrm{sign}(\Gamma_{1,n}),\ldots,\mathrm{sign}(\Gamma_{K,n}))^\top$.

For $s\in\R$, define the FPR $\alpha_n(s):=\mathbb P(S_n>s\mid H_0)$ and
the TPR $\pi_n(s):=\inf_{\bm m\in \{\pm 1\}^K}\mathbb P(S_n>s\mid H_1,\bm m)$ for~\eqref{eq:hywithnoD_Ik_n}, and define
the FPR $\alpha_{\rm orc}(s):=\mathbb P(S>s\mid H_0)$ and the TPR $\pi_{\rm orc}(s):=\inf_{\bm m\in \{\pm 1\}^K}\mathbb P(S>s\mid H_1,\bm m)$ for~\eqref{eq:hywithnoD_Ik}.

Define $S_{\mathcal D,n}:=\max_{\bm m\in\mathcal D}\langle \bm m,\bm \Gamma_n\rangle$. With Theorem~\ref{thm:ikdist_n}, the watermark detection problem~\eqref{eq:hpdetectD} can be reduced to the following one:
{
\vspace{-32pt}

\begin{equation}\label{eq:hywithD_Ik_n}
H_0: \bm \Gamma_n \sim \mathcal N(\bm 0,\Sigma_{n})
\quad H_{1'}: \bm \Gamma_n \sim \mathcal N(G_n\bm m,\Sigma_{n}) \text{ for some message $\bm m \in\mathcal D$}.
\end{equation}

\vspace{-20pt}
}
Following the detection and decoding rules derived in Theorem~\ref{thm:dict_LRT}, reject $H_0$ when $S_{\mathcal D,n}>t$ for a threshold $t$, and decode with $\widehat{\bm m}^{\mathcal D}_n: =\arg\max_{\bm m\in\mathcal D}\ \langle \bm m,\bm \Gamma_n\rangle$. 

For any threshold $t\in\R$, define the FPR $\alpha_{\mathcal D,n}(t):=\mathbb P(S_{\mathcal D,n}>t\mid H_0)$ and the TPR
$\pi_{\mathcal D,n}(t):=\inf_{\bm m\in\mathcal D}\mathbb P(S_{\mathcal D,n}>t\mid H_1,\bm m)$ for~\eqref{eq:hywithD_Ik_n}, and 
define the FPR $\alpha_{\mathcal D,{\rm orc}}(t):=\mathbb P(S_{\mathcal D}>t\mid H_0)$ and the TPR
$\pi_{\mathcal D,{\rm orc}}(t):=\inf_{\bm m\in\mathcal D}\mathbb P(S_{\mathcal D}>t\mid H_1,\bm m)$ for~\eqref{eq:hywithD_Ik}.

\begin{theorem}[Convergence of FPR, TPR, and bit accuracy]\label{thm:TV_transfer_scores}
Suppose that the same conditions of Theorem~\ref{thm:finite_sample_geometry} hold and $\sigma_\epsilon>0$. For any $\delta\in(0,1)$,  let $ \varepsilon_n(\delta)$ be as defined in Theorem~\ref{thm:finite_sample_geometry}.
Then there exists a constant $\tau_1>0$ that does not depend on $n$, such that if  $\varepsilon_n(\delta)\leq \tau_1$, the following hold with probability at least $1-\delta$
over the training sample $\{\bm x_i\}^n_{i=1}$:
\begin{enumerate}[label=(\roman*),leftmargin=2em]
\item \textbf{FPR and TPR with $S_n$.}  For any $s\in\R$,
$|\alpha_n(s)-\alpha_{\rm orc}(s)|\lesssim \varepsilon_n^{1/4}(\delta)+\varepsilon_n^{1/2}(\delta)+\varepsilon_n(\delta)$ and
$|\pi_n(s)-\pi_{\rm orc}(s)|\lesssim \varepsilon_n^{1/4}(\delta)+\varepsilon_n^{1/2}(\delta)+\varepsilon_n(\delta)$.
\item \textbf{FPR and TPR with $S_{\mathcal D, n}$.} Fix a dictionary $\mathcal D\subset\{\pm1\}^K$. 
For any $t\in\R$, $|\alpha_{\mathcal D,n}(t)-\alpha_{\mathcal D,{\rm orc}}(t)|\lesssim   \varepsilon_n^{1 / 4}(\delta)+\varepsilon_n^{1 / 2}(\delta)+\varepsilon_n(\delta)  $ and
$|\pi_{\mathcal D,n}(t)-\pi_{\mathcal D,{\rm orc}}(t)|\lesssim  \varepsilon_n^{1 / 4}(\delta)+\varepsilon_n^{1 / 2}(\delta)+\varepsilon_n(\delta).$
\item \textbf{Bit accuracy.} $\underset{\bm m\in\{\pm1\}^K}{\sup}
\Big|
\E[\textsc{ba}(\widehat{\bm m}^{\mathrm{sign}}_{n},\bm m)\mid H_1,\bm m]
-
\E[\textsc{ba}(\widehat{\bm m}^{\mathrm{sign}},\bm m)\mid H_1,\bm m]
\Big|
\lesssim
\varepsilon_n^{1/4}(\delta)+\varepsilon_n^{1/2}(\delta)+\varepsilon_n(\delta)$ and $\underset{\bm m\in\mathcal D}{\sup}
\Big|
\E[\textsc{ba}(\widehat{\bm m}^{\mathcal D}_n,\bm m)\mid H_1,\bm m]
-
\E[\textsc{ba}(\widehat{\bm m}^{\mathcal D},\bm m)\mid H_1,\bm m]
\Big|
\lesssim
\varepsilon_n^{1/4}(\delta)+\varepsilon_n^{1/2}(\delta)+\varepsilon_n(\delta).$
\end{enumerate}

\end{theorem}

\vspace{-10pt}
\section{Experiments}\label{sec:experiments}

In this section, we evaluate the performance of ADD through experiments on real image datasets. In Section~\ref{sec:perf}, we compare the performance of ADD with representative watermarking methods under a wide range of image distortions. In Section~\ref{sec:compute}, we demonstrate the computational advantage of ADD. In Section~\ref{ssubsec:gen_semantics}, we demonstrate the generalizability of ADD to other datasets. In Section~\ref{sec:trade-offs}, we present and discuss the empirical trade-offs from hyperparameters $\beta$ and $n$.

\textbf{Setup.} The training data is sampled from the train split of MS-COCO~\citep{mscoco} dataset, which is one of the most widely used large-scale benchmarks in computer vision and contains natural images depicting a wide range of real-world scenes and objects. Its train split contains about $118{,}000$ images, and the full dataset includes over $330{,}000$ images with annotations for $80$ object categories and more than $1.5$ million object instances.  All images are resized to a fixed resolution of \(256\times256\) pixels and processed as RGB images. We set $K=48$ for the multi-bit message, which already represents a challenging regime to the best of our knowledge. Existing watermarking methods often experience substantial degradation in decoding performance at $K=48$, while ADD is not restricted to this value of $K$. The metrics for the three goals of watermarking as discussed in Section~\ref{sec:formulation} are as follows: 1)\textbf{Quality} is evaluated by the Peak Signal-to-Noise Ratio (PSNR) (a widely used metric for image watermarking~\citep{coxDigitalWatermarkingSteganography2007}) between the original image \(\bm{x}\) and the watermarked image \(\tilde{\bm{x}}\), defined as
$\text{PSNR} = 10 \log_{10} \left( \frac{\text{MAX}^2}{\text{MSE}} \right),$
where $\text{MAX}$ denotes the maximum possible pixel value of the image (255 for an 8-bit image), and
$\text{MSE} := \frac{1}{C\cdot H\cdot W}\sum_{c=1}^{C}\sum_{h=1}^{H}\sum_{w=1}^{W}
\left( \bm{x}_{c,h,w}-\tilde{\bm{x}}_{c,h,w}\right)^2,$ where $C$, $H$, and $W$ denote the number of channels, image height, and image width, respectively. 2)\textbf{Identifiability} is evaluated by the area under the receiver operating characteristic curve (AUROC) for  detection performance, and bit accuracy for decoding performance. 3)\textbf{Resilience} is evaluated by the detection and decoding performance under a wide range of image distortions introduced in~\cite{anWAVESBenchmarkingRobustness2024}. See details of these image distortions in Section~E of the supplementary material. For training with Algorithm~\ref{alg:train}, we set $\beta =1000$ (note that in Algorithm~\ref{alg:train} we scaled down the penalty term by $D$ so this corresponds to $\beta = 1000/(256^2\times3) \approx 0.005$ in Section~\ref{sec:theory}) and $n = 2000$. The remaining training details are provided in Section~F of the supplementary material.

\textbf{Baselines.} The representative baseline methods we considered include: DwtDct~\citep{al-hajCombinedDWTDCTDigital2007}, a traditional frequency-based method deployed by a popular generative model Stable Diffusion~\citep{rombachHighResolutionImageSynthesis2022}; HiDDeN~\citep{zhuHiddenHidingData2018}, a widely used deep learning-based method; SSL~\citep{fernandezWatermarkingImagesSelfsupervised2022}, a watermarking method that optimizes the watermark for each image during embedding and typically achieves strong performance, while decoding the multi-bit message with inner products, which is similar to our decoding rule. Together, these baselines cover both traditional frequency-based and deep learning-based watermarking methods, and represent competitive approaches commonly used in prior watermarking studies.

\vspace{-10pt}
\subsection{Competitive performance of ADD}\label{sec:perf}

\begin{figure}[htb]
    \centering
    \includegraphics[width=1\textwidth]{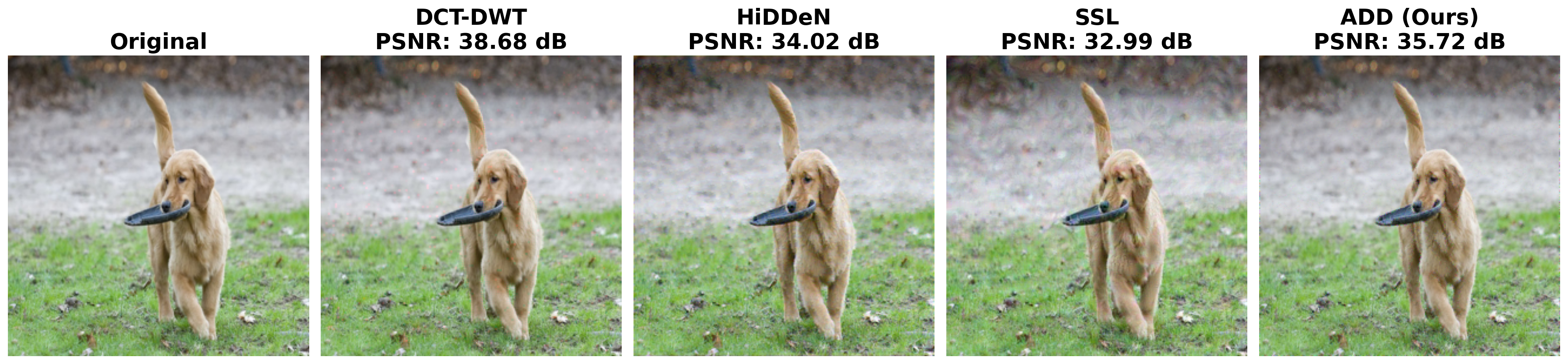}
    \caption[comparison]{\textbf{Qualitative comparison of image quality across watermarking methods.}
    All methods are evaluated on the same image, with PSNR values reported. Our method is visually close to the original image, as the rest of the methods do. A visual comparison of the magnified pixel-wise differences is in  Section~G of the supplementary material.}
    \label{fig:method_fidelity}
\end{figure}

For multi-bit watermarking, decoding performance is the primary quantity of interest. As shown in Table~\ref{tab:wm_decode_test2017}, our method achieves the best decoding performance (bit accuracy) under all distortion settings, while preserving a comparable quality (PSNR) with others. A qualitative comparison of image quality is shown in Figure~\ref{fig:method_fidelity}, showing that our method maintains visual fidelity to the original image comparable to the competing methods.

\begin{table*}[htb]
\centering
\small
\setlength{\tabcolsep}{5pt}
\renewcommand{\arraystretch}{1.15}
\caption[Main decoding result]{\textbf{Watermark decoding results.} All methods are evaluated on the same $1000$ randomly sampled image--message pairs from the MS-COCO test split.
PSNR (dB, higher is better) is reported as mean $\pm$ standard error. Bit accuracy (\%, higher is better) is reported in under each distortion setting. The \emph{Average} column reports the mean bit accuracy across all distortion settings.}
\label{tab:wm_decode_test2017}
\resizebox{\textwidth}{!}{
\begin{tabular}{l c ccccccccc c}
\toprule
\textbf{Method} & \textbf{PSNR} &
\myrothead{\textbf{None}} &
\myrothead{\textbf{Gaussian Blur}} &
\myrothead{\textbf{JPEG} }&
\myrothead{\textbf{Brightness}} &
\myrothead{\textbf{Contrast}} &
\myrothead{\textbf{Gaussian Noise}} &
\myrothead{\textbf{Rotation}} &
\myrothead{\textbf{Crop}} &
\myrothead{\textbf{Random Erase} }&
\myrothead{\textbf{Average}} \\
\midrule
\textbf{DwtDct} & $\mathbf{37.22} \pm 0.08$ &
89.2 & 50.8 & 50.5 & 46.7 & 54.1 & 50.0 & 51.3 & 69.5 & 76.6 & 59.8 \\
\textbf{HiDDeN} & $32.88 \pm 0.05$ &
99.7 & 57.6 & 91.2 & 98.2 & 99.6 & 50.5 & 49.2 & 98.5 & 98.1 & 82.5 \\
\textbf{SSL} & $33.09 \pm 0.00$ &
100.0 & 93.8 & 88.4 & 88.6 & 91.8 & 54.7 & 96.6 & 80.7 & 69.9 & 84.9 \\
\midrule
\rowcolor{gray!12}
\textbf{ADD (Ours)} & $32.36 \pm 0.06$ &
\textbf{100.0} & \textbf{98.1} & \textbf{98.6} & \textbf{99.6} & \textbf{99.9} &
\textbf{98.8} & \textbf{99.8} & \textbf{99.9} & \textbf{99.9} & \textbf{99.4} \\
\bottomrule
\end{tabular}
}
\end{table*}

Watermark detection is also evaluated as part of the performance. Among competing methods considered in this paper, there is no implementation that can detect an image without the assistance of a dictionary of embedded messages. Therefore, for these methods, detection is performed via a matching-bit test: given a decoded message $\bm m$, declare an image as watermarked when the maximum number of matched bits with entries in the message dictionary $\mathcal D$ exceeds a threshold $\gamma$, i.e., when $\max_{\bm m\in\mathcal D}\sum_{k=1}^K \mathbbm{1}\{\hat m_k=m_k\}>\gamma$. Our method, on the other hand, can do detection either with or without a dictionary of embedded messages.

The results in Table~\ref{tab:wm_detect_coco10k} show that our method achieves the strongest overall detection performance, as reflected by the highest average AUROC across distortions. 
Even without access to a message dictionary, our method substantially outperforms competing methods on average. 
When a dictionary $\mathcal D$ is available, our method further improves performance and yields the best overall detection results. The results of the receiver operating characteristic curve (ROC) are in Section~H of the supplementary material.

\begin{table*}[htb]
\centering
\small
\setlength{\tabcolsep}{5pt}
\renewcommand{\arraystretch}{1.15}
\caption[Watermark detection results]{\textbf{Watermark detection results.}
All methods are evaluated on the same $10{,}000$ randomly sampled image--message pairs from the MS-COCO test split.
PSNR (dB, higher is better) is reported as mean.
AUROC (\%, higher is better) is reported under each distortion setting. The \emph{Average} column reports the mean AUROC across all distortion settings.
DwtDct/HiDDeN/SSL use \emph{dictionary-based} detection with dictionary $\mathcal{D}$ ($d_{\min}=6$) consisting of all $10{,}000$ embedded 48-bit messages (threshold $\gamma\in\{0,\ldots,48\}$ on best-match score).
ADD (w/o $\mathcal{D}$) uses the dictionary-agnostic statistic $S$ (threshold sweep on $S$).
ADD (w/ $\mathcal{D}$) uses the dictionary-dependent statistic $S_{\mathcal D}$ (threshold sweep on $S_{\mathcal D}$).}
\label{tab:wm_detect_coco10k}
\resizebox{\textwidth}{!}{
\begin{tabular}{lccccccccccc}
\toprule
\textbf{Method} & \textbf{PSNR} &
\myrothead{\textbf{None}} &
\myrothead{\textbf{Gaussian Blur}} &
\myrothead{\textbf{JPEG}} &
\myrothead{\textbf{Brightness}} &
\myrothead{\textbf{Contrast}} &
\myrothead{\textbf{Gaussian Noise}} &
\myrothead{\textbf{Rotation}} &
\myrothead{\textbf{Crop}} &
\myrothead{\textbf{Random Erase}} &
\myrothead{\textbf{Average}} \\
\midrule
\textbf{DwtDct} & \textbf{37.05} &
88.0 & 49.8 & 50.0 & 50.5 & 52.7 & 49.8 & 49.8 & 54.9 & 69.0 & 57.2 \\
\textbf{HiDDeN} & 32.82 &
\textbf{100.0} & 48.4 & 97.6 & 99.2 & 99.9 & 50.0 & 49.5 & 99.9 & 99.3 & 82.6 \\
\textbf{SSL} & 33.09 &
\textbf{100.0} & \textbf{99.6} & 95.9 & 92.1 & 95.8 & 50.4 & 99.9 & 81.4 & 55.3 & 85.6 \\
\midrule
\rowcolor{gray!12}
\textbf{ADD (w/o $\mathcal{D}$)} & &
\textbf{100.0} & 94.5 & 95.0 & 99.5 & \textbf{100.0} & \textbf{100.0} & \textbf{100.0} & \textbf{100.0} & \textbf{100.0} & 98.8 \\
\rowcolor{gray!12}
\textbf{ADD  (w/  $\mathcal{D}$)} & \multirow{-2}{*}{\cellcolor{gray!12}32.27} &
\textbf{100.0} & 98.5 & \textbf{98.3} & \textbf{99.7} & \textbf{100.0} &
\textbf{100.0} & \textbf{100.0} & \textbf{100.0} & \textbf{100.0} & \textbf{99.6} \\
\bottomrule
\end{tabular}
}
\end{table*}

\vspace{-10pt}
\subsection{Computational advantage of ADD}\label{sec:compute}

A side benefit of our method is its computational efficiency. As reported in Table~\ref{tab:runtime_coco_a100}, our approach is at least $2\times$ faster at the embedding stage and $7.4\times$ faster at the decoding stage than competing methods. This improvement arises from the simplicity of the underlying operations: embedding only requires linear addition of the watermark, while decoding reduces to computing inner products. In contrast, competing methods with comparable performance typically rely on more sophisticated computations or neural network–based processing, resulting in substantially higher runtime.

\begin{table}[htb]
\centering
\small
\setlength{\tabcolsep}{10pt}
\caption[Runtime comparison]{\textbf{Runtime comparison of watermark embedding and decoding.} We report average time per image (ms/image) $\pm$ standard error for the watermark \emph{embedding} step (embedding a multi-bit watermark $\boldsymbol{m}$ into an image) and the \emph{decoding} step (recovering the embedded bits $\hat{\boldsymbol{m}}$ from the watermarked image). All methods were evaluated on the same 1{,}000 images from the MS-COCO test split using a single NVIDIA A100 GPU. Standard errors are computed across processing batches (batch size 64), treating each batch's per-image time as one observation.}
\label{tab:runtime_coco_a100}
\begin{tabular}{lcc}
\toprule
\textbf{Method} & \textbf{Embedding (ms/img)} & \textbf{Decoding (ms/img)} \\
\midrule
\textbf{DwtDct}        & $8.37 \pm 0.01$  & $5.41 \pm 0.01$ \\
\textbf{HiDDeN}        & $1.54 \pm 0.01$  & $1.41 \pm 0.01$ \\
\textbf{SSL}           & $546.50 \pm 2.37$ & $6.50 \pm 0.00$ \\
\midrule
\rowcolor{gray!12}
\textbf{ADD (Ours)}    & $\mathbf{0.76 \pm 0.01}$ & $\mathbf{0.19 \pm 0.00}$ \\
\bottomrule
\end{tabular}
\end{table}

\vspace{-10pt}
\subsection{Generalizability to other datasets}
\label{ssubsec:gen_semantics}

Here we evaluate our watermark $\bm w_{1:K}$ (trained with Algorithm~\ref{alg:train} \emph{once} using $n=2000$ images from MS-COCO) on three other popular datasets, including ImageNet~\citep{imagenet}, CIFAR-10 and CIFAR-100~\citep{cifar}.

Overall, our watermark trained only on MS-COCO generalizes well to multiple unseen domains without retraining. Table~\ref{tab:expA_summary} summarizes the result on MS-COCO (in-domain) and on out-of-domain datasets
(ImageNet, CIFAR-10, and CIFAR-100), each with $1000$ images.
Across all datasets, decoding remains nearly perfect ($\ge 99.27\%$ bit accuracy) and detection remains good (AUROC $\ge 0.9893$).
Note that on CIFAR-10 and CIFAR-100 ADD achieves higher PSNR due to up-sampling from $32\times32$ to $256\times 256$, which yields smoother images.

\begin{table}[htb]
\centering
\caption[generalization]{\textbf{Cross-dataset generalization.} Our watermark is trained on MS-COCO only and evaluated on multiple test domains, each with 1000 images.
We report PSNR (mean $\pm$ standard error), average bit accuracy, and average AUROC. Bit accuracy and AUROC are averaged over the performance under each distortion setting. See detailed per-distortion result in Section~I of the supplementary material.}
\label{tab:expA_summary}
\begin{tabular}{lccc}
\toprule
\textbf{Test domain} & \textbf{PSNR (dB)} & \textbf{Avg Bit Accuracy} & \textbf{Avg AUROC} \\
\midrule
MS-COCO (in-domain) & $32.36 \pm 0.06$ & $99.37\%$ & $0.9927$ \\
ImageNet            & $32.21 \pm 0.07$ & $99.33\%$ & $0.9893$ \\
CIFAR-10            & $36.16 \pm 0.08$ & $99.37\%$ & $0.9987$ \\
CIFAR-100           & $35.75 \pm 0.09$ & $99.27\%$ & $0.9978$ \\
\bottomrule
\end{tabular}
\end{table}

\vspace{-10pt}
\subsection{Empirical trade-offs from $\beta$ and $n$}\label{sec:trade-offs}

\begin{figure}[htb]
    \centering
    \begin{subfigure}[b]{0.49\textwidth}
        \centering
        \includegraphics[width=\textwidth]{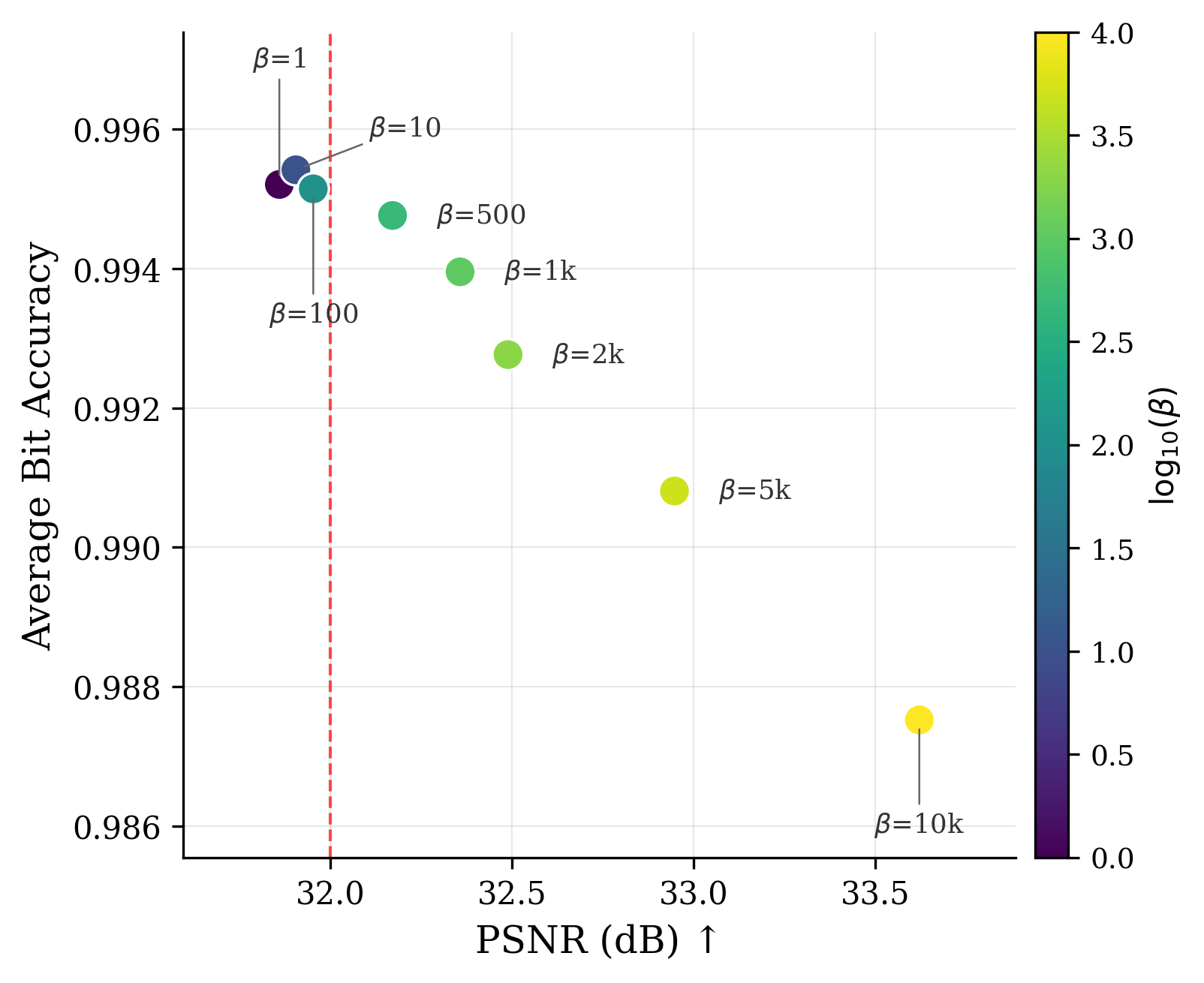}
        \subcaption{Varying $\beta$.}
        \label{fig:tradeoff_beta}
    \end{subfigure}
    \hfill
    \begin{subfigure}[b]{0.49\textwidth}
        \centering
        \includegraphics[width=\textwidth]{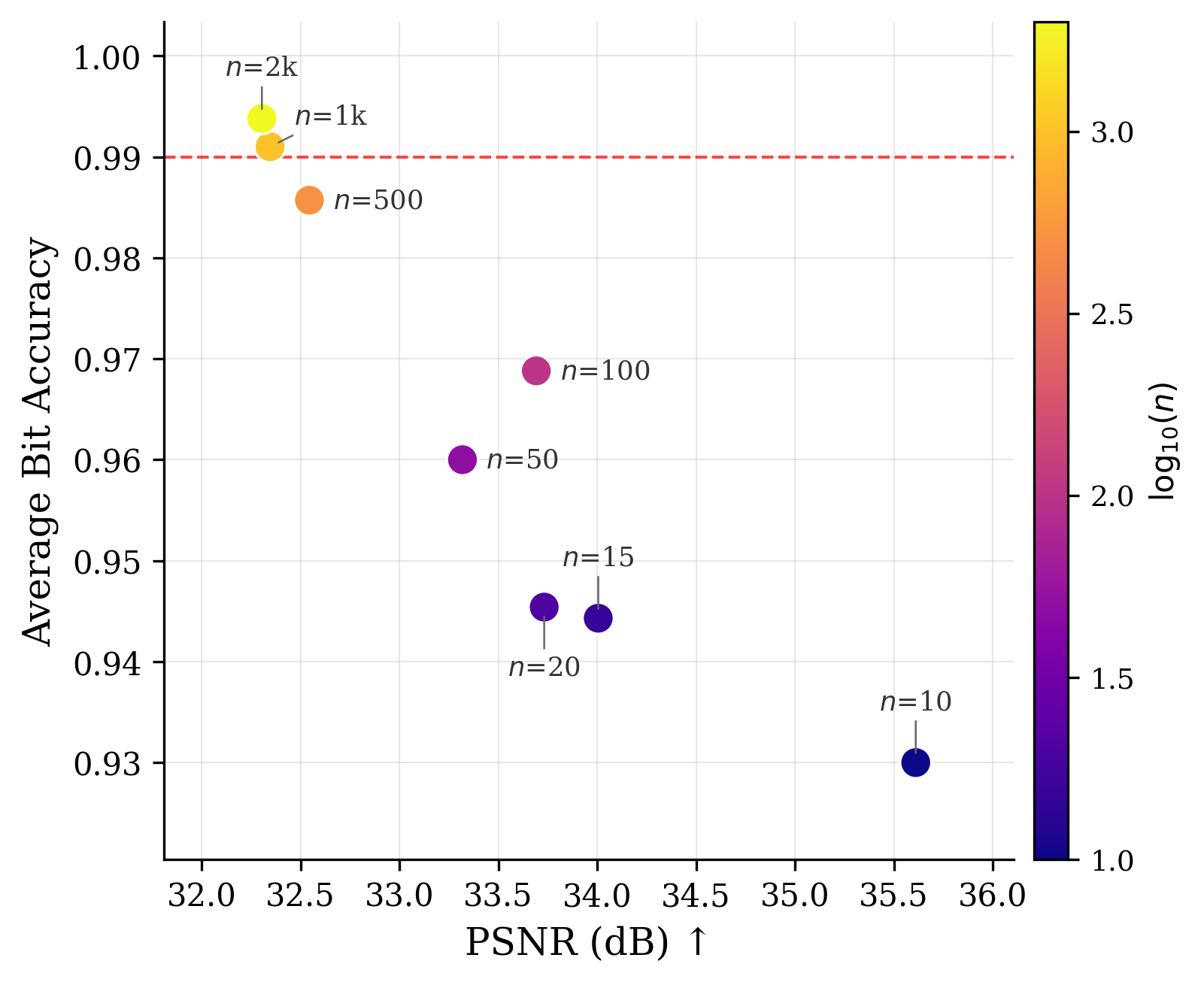}
        \subcaption{Varying sample size $n$.}
        \label{fig:tradeoff_n}
    \end{subfigure}
    \caption[trade-off]{\textbf{Empirical trade-offs from $\beta$ and $n$.} Trade-off between PSNR and average bit accuracy from (a) different $\beta$ values and (b) different training sample sizes $n$. In (a), $n=2000$ and $\beta\in\{1,10,100,500,1000,5000,10000\}$. In (b), $\beta=1000$ and $n\in\{10,15,20,50,100,500,1000,2000\}$. The dashed red lines indicate acceptable thresholds chosen to reflect practical deployment requirements: PSNR $=32$ dB in (a) and
average bit accuracy $=0.99$ in (b). The rest of the training setup is the same.}
    \label{fig:comparison}
\end{figure}

Figure~\ref{fig:comparison} illustrates how the empirical performance of ADD varies with the regularization parameter $\beta$ and the training sample size $n$.

Panel~(a) in Figure~\ref{fig:comparison} shows a clear trade-off from $\beta$ between image quality and decoding performance. As $\beta$ increases, PSNR increases while the average bit accuracy decreases. This is well aligned with the practical implication in Section~\ref{sec:ddpop}: a larger $\beta$ enforces stronger regularization, which leads to a smaller watermark magnitude $r_\star$, and hence better image quality but weaker decoding performance.  

Panel~(b) in Figure~\ref{fig:comparison} shows that as $n$ increases, the average bit accuracy improves substantially, while PSNR remains relatively stable when $n\geq 500$. This is consistent with our finite-sample theory: larger $n$ reduces the discrepancy between the finite-sample and population objectives, so the learned watermark is closer to its population counterpart. When $n$ is small, this discrepancy is larger and the learned watermark is less reliable.

\vspace{-10pt}
\section{Conclusion}\label{sec:conclusion}

We propose ADD, a multi-bit image watermarking method that learns an additive watermark, embeds a $K$-bit message by linear combination, and performs detection and decoding using only inner products with the stored watermark. On MS-COCO, ADD achieves near-perfect decoding under common distortions while maintaining competitive visual quality, outperforming competing methods and offering substantially faster embedding and decoding due to its simple structure. We further provide a theoretical explanation for why ADD works. Under a low-dimensional subspace model for images, we show that the population objective yields watermark that is orthogonal to the image subspace and mutually orthogonal, which leads naturally to a generalized likelihood ratio test for detection and a corresponding decoding rule. We further establish that, for the finite-sample watermark, the FPRs, TPRs, and bit accuracy under the derived detection and decoding rules converge to their population counterparts as the training sample size grows.

We highlight two directions for future work. First, a natural next step would be to extend ADD for other modalities such as video, audio, and text, which will require modality-specific adjustments. Second, this paper assumes that the images are watermarked by only one watermarking mechanism. In reality, different entities may deploy different watermarking mechanisms, which will require a centralized allocation and verification of messages. We discuss our vision on this in Section~B of the supplementary material, which points out a future direction for scaling provenance mechanisms and enabling accountable use of generative media in practical applications.

\vspace{-10pt}
\section*{Use of Generative AI Tools}

During the preparation of this manuscript, the authors used ChatGPT-5.2 (OpenAI) and AgentLab (MorphMind) for language improvement and figure design, and Claude Opus 4.6 (Anthropic) for coding assistance. These tools were used only to improve clarity of writing, assist with programming, and support figure preparation. The authors reviewed and edited all outputs and take full responsibility for the content of this manuscript.

\vspace{-20pt}
\bibliographystyle{agsm}
\bibliography{refs}

\end{document}